\documentclass[journal]{IEEEtran}

\usepackage{multirow}

\ifCLASSINFOpdf
  \usepackage[pdftex]{graphicx}
\else
\fi

\usepackage{amsmath}

\usepackage{array}

\usepackage{cite}
\usepackage{graphicx}
\usepackage{bm}
\usepackage{caption}
\usepackage{hyperref}
\usepackage{amssymb}
\usepackage{color,xcolor}
\usepackage{subfigure}

\usepackage{algorithm}
\usepackage{algorithmicx}
\usepackage{subfigure}
\usepackage{algpseudocode}

\usepackage{ragged2e}

\begin{document}
%
\title{Multimodal Information Bottleneck: Learning Minimal Sufficient Unimodal and Multimodal Representations}
%
%
%

\author{Sijie Mai$^1$,
        Ying Zeng$^1$,
        Haifeng Hu
\IEEEcompsocitemizethanks{\IEEEcompsocthanksitem Haifeng Hu (corresponding author)
 is with the School of Electronics and Information Technology, Sun Yat-sen University,
China.\protect\\
E-mail: huhaif@mail.sysu.edu.cn\protect\\
This work was supported by the National Natural Science Foundation of China under Grant 62076262.
\IEEEcompsocthanksitem $^1$These authors contribute equally.
}
\thanks{ }}

\markboth{IEEE TRANSACTIONS ON MULTIMEDIA}%
{Shell \MakeLowercase{\textit{et al.}}: Bare Demo of IEEEtran.cls for IEEE Journals}

\IEEEtitleabstractindextext{%
\begin{abstract}
\justifying
Learning effective joint embedding for cross-modal data has always been a focus in the field of multimodal machine learning. We argue that during multimodal fusion, the generated multimodal embedding may be redundant, and the discriminative unimodal information may be ignored, which often interferes with accurate prediction and leads to a higher risk of overfitting. Moreover, unimodal representations also contain noisy information that negatively influences the learning of cross-modal dynamics. To this end, we introduce the multimodal information bottleneck (MIB), aiming to learn a powerful and sufficient multimodal representation that is free of redundancy and to filter out noisy information in unimodal representations. Specifically, inheriting from the general information bottleneck (IB), MIB aims to learn the minimal sufficient representation for a given task by maximizing the mutual information between the representation and the target and simultaneously constraining the mutual information between the representation and the input data. Different from general IB, our MIB regularizes both the multimodal and unimodal representations, which is a comprehensive and flexible framework that is compatible with any fusion methods. We develop three MIB variants, namely, early-fusion MIB, late-fusion MIB, and complete MIB, to focus on different perspectives of information constraints. Experimental results suggest that the proposed method reaches state-of-the-art performance on the tasks of multimodal sentiment analysis and multimodal emotion recognition across three widely used datasets. The codes are available at \url{https://github.com/TmacMai/Multimodal-Information-Bottleneck}.

\end{abstract}

}

\maketitle

\IEEEdisplaynontitleabstractindextext

%
\IEEEpeerreviewmaketitle

\section{\textbf{Introduction}}\label{sec:Introduction}
\IEEEPARstart{W}ith the growing trend of sharing lives on social media, videos that express human opinion are easily accessed, which is worthwhile for the study of understanding the implicit meanings. In those videos, people express their views and emotions through multiple modalities. Apart from spoken language, facial expressions and acoustic clues are also informative during communication between humans or between humans and machines \cite{RMFN,8269806,TASLP}. The three modalities are often complementary and interact with each other. Consequently, a multimodal learning architecture is necessary for a comprehensive interpretation of human language.

Previous works have tended to design complex fusion strategies to learn multimodal embedding with unimodal features, such as tensor fusion \cite{Zadeh2017Tensor} and transformer-based fusion \cite{transformer,MULT}. Though effective, these methods suffer from high complexity, and the generated high-dimensional multimodal embedding is inevitably redundant, which leads to a higher risk of overfitting. Ideally, multimodal embedding should (1) contain the maximum information that is necessary for correct prediction and (2) contain the least information that is irrelevant to the prediction.
However, it is not straightforward to learn such perfect embedding.
We argue that the multimodal embedding generated by the complex fusion network may be redundant such that the discriminative unimodal information is ignored. For example, Zadeh \textit{et al.} \cite{Zadeh2017Tensor} used an outer product to generate a high-order multimodal tensor, which produces redundant representations that may overwrite the useful unimodal information during prediction.

Moreover, unimodal representations inevitably contain noisy information, which may negatively influence correct prediction. In the field of multimodal sentiment analysis and emotion recognition, previous studies \cite{TASLP,HFFN} have demonstrated that the language modality is always dominant, while the visual and acoustic modalities are more like auxiliary modalities. These nonlexical modalities often contain noise that may interfere with the learning of cross-modal interactions and influence the quality of the multimodal representation.

How can we ensure that the learned latent representation is sufficient for correct inference and simultaneously free of redundancy and noise? In this paper, drawing inspiration from the information bottleneck \cite{ib,vib,tmm_ib}, we propose a multimodal information bottleneck (MIB) to learn such unimodal and multimodal representations. The information bottleneck (IB) is based on mutual information (MI), aiming to maximize the MI between the encoded representations and the labels and simultaneously minimize the MI between the encoded representation and the input.  IB is able to find the concise and compressed representation of the input, taking into account the complexity of a signal from an information theory perspective\cite{ib}. By applying the IB principle, the model can learn to filter out noisy and redundant information that might interfere with the prediction and thus obtain minimal sufficient representation for prediction. In our work, IB is used to extract concise representation from cross-modal embedding that discards the irrelevant information to the target.
A multimodal sample carries considerable information, and the multimodal embedding generated by complicated fusion methods suffers from high complexity and high dimensionality, which easily leads to redundancy. Therefore, extracting concise representations for multimodal learning is significant.

Specifically, to reduce the redundant information in the generated multimodal representation, we propose an early-fusion MIB (E-MIB). The early fusion here denotes that we first generate the primary multimodal representation via fusion and then, based on it, learn the minimal sufficient multimodal representation via the IB principle (the early fusion here is different from the commonly considered feature concatenation). First, we extract and fuse the high-level unimodal representations of the three modalities to obtain the primary joint multimodal representation, where the fusion method is flexible.
Then, we feed the primary joint multimodal representation to the information bottleneck, based on which the multimodal representation is distilled and compressed. Ideally, the compressed joint representation only contains information that is relevant to the task, and noisy or redundant information is filtered out. The compressed multimodal representation is therefore free of redundancy, and the risk of overfitting can be reduced.

Moreover, to filter out noise that is irrelevant to the prediction in the unimodal representation, we propose a late-fusion MIB (L-MIB). The late fusion here denotes that we first leverage the IB principle to learn the minimal sufficient unimodal representations and based on them generate the multimodal representation via fusion (the late fusion here is different from the commonly considered late fusion, i.e. voting based on unimodal predictions). By this means, the extracted concise unimodal representation is free of noise, and the generated multimodal representation can be more discriminative. In addition, by applying the IB principle, the distribution gap between modalities can be reduced, such that they can be better fused.

Last but not least, to comprehensively inject the IB principle into multimodal systems, E-MIB and L-MIB are combined to construct the complete MIB framework (C-MIB). C-MIB takes into account filtering out noise in the unimodal representations as well as learning a minimal sufficient multimodal representation, which is expected to be more comprehensive. We compare the performance of L-MIB, E-MIB and C-MIB in Section~\ref{sec:fusion_e}. Note that our MIB framework is highly flexible, which is compatible with any fusion mechanisms to provide higher expressive power. The comparison of fusion mechanisms is also shown in Section~\ref{sec:fusion_e}.

In conclusion, we propose a new architecture named the multimodal information bottleneck (MIB) to address the multimodal learning problem. The contributions are listed below:
\begin{itemize}
  \item We propose a novel architecture to model multimodal signals. We innovate to inject the IB principle into multimodal learning to filter out noisy information and reduce redundancy. To the best of our knowledge, we are the first to leverage the IB principle to learn both minimal sufficient unimodal and multimodal representations in multimodal sentiment analysis and emotion recognition.
  \item E-MIB and L-MIB are proposed to learn minimal sufficient multimodal representation and filter out the noise in unimodal representations, respectively. Combining the advantages of E-MIB and L-MIB, C-MIB is proposed to comprehensively learn more favorable unimodal and multimodal representations. Our MIB framework is generalized to be applied with commonly used fusion strategies to provide higher expressive power.
  \item The MIB variants outperform other methods on multiple datasets. In addition, the visualization experiments demonstrate that our MIB enables the multimodal representations to be more discriminative. Moreover, our method can reach state-of-the-art performance even with very simple fusion methods.
\end{itemize}


\section{\textbf{Related Work}}\label{sec:Related Work}
\label{sec:format}
\subsection{Multimodal sentiment analysis and emotion recognition}
Multimodal sentiment analysis and emotion recognition  have attracted significant research attention recently \cite{8269806}. Previous publications focus on learning various fusion strategies to explore inter-modal dynamics \cite{tmm_urfn,Poria2017A,Poria2017Convolutional,Pang2015Deep}. One of the simplest ways is to concatenate features at input feature level, which demonstrates improvement over single modality \cite{Wollmer2013YouTube,Rozgic2012Ensemble,Poria2017Convolutional}. In contrast, a large number of prior works firstly infer prediction according to each modality and combine the predictions from all modalities using some voting mechanisms \cite{Wu2010Emotion,Nojavanasghari2016Deep,Personality,Zadeh2016MOSI}. Nevertheless, these two types of fusion methods cannot explicitly model cross-modal interactions \cite{Zadeh2017Tensor}.

Consequently, more advanced fusion strategies have been proposed in the past few years. Specifically, performing tensor fusion has received much attention \cite{T2FN, Liu2018Efficient,LMFN}. Tensor Fusion Network (TFN) \cite{Zadeh2017Tensor}  adopts outer product to learn joint representation of three modalities. However, the generated multimodal representation is  high-dimensional and of high-complexity, which contains redundant information and may suffer from overfitting. Furthermore, modality-translation methods such as Multimodal Cyclic Translation Network (MCTN) \cite{MCTN} and Multimodal Transformer  \cite{MULT} aim at learning joint representation by translating source modality into the target one.
More recently, graph-based fusion methods have emerged, which can effectively extract interaction across time series or modalities \cite{ARGF,MOSEI,mai2020analyzing}. For example, Graph-MFN \cite{MOSEI} uses a dynamic fusion graph to fuse features, and Mai \textit{et al.} \cite{ARGF} conduct fusion using a graph fusion network which regards each modality as one node.
Furthermore, using cross-modal attention mechanism to explore inter-modal dynamics are also very popular \cite{MTL,Multilogue-Net}. To avoid sarcasm and ambiguity, many methods align the three modalities in the time dimension  and then fuse the unimodal features  at word level \cite{TASLP,RAVEN,Zadeh2018Multi,RMFN}. For instance, Recurrent Attended Variation Embedding Network (RAVEN) \cite{RAVEN} models multimodal language by shifting word representations based on audio and vision clues at word level. Moreover, aiming to address the interpretable problem in multimodal fusion, quantum-inspired fusion \cite{Quantum} and capsule-based fusion \cite{MRM} are emerged.

More recently, with the success of Bidirectional Encoder Representations from Transformers (BERT) \cite{BERT}, fine-tuning BERT has obtained new state-of-the-art results on many natural language processing (NLP) tasks. The original BERT is a pretrained language model, which only accepts lexical information.
MAG-BERT \cite{MAG-BERT} proposes an attachment, namely Multimodal Adaptative Gate (MAG), to allow BERT and XL-Net \cite{XLNet} to accept multimodal nonverbal data during fine-tuning. Cross-Modal BERT (CM-BERT) \cite{CM-BERT} relies on the interaction of text and audio modality to fine-tune the pretrained BERT model, where a masked multimodal attention is designed to dynamically adjust the weight of words by combining the combination of text and audio modality.

Different from the above works, our MIB focuses on learning the minimal sufficient representation instead of designing sophisticated fusion strategies, which is flexible as it can integrate any fusion mechanisms. The proposed MIB is elegant and effective, which can  filter out noisy information and reduce the redundancy of the generated representations. To the best of our knowledge, few prior works have employed the information bottleneck principle to explore both minimal sufficient multimodal and unimodal representations, and this paper offers a solution to explore the feasibility of information bottleneck on multimodal learning tasks.

\subsection{Information Bottleneck}

Information bottleneck (IB) is originally proposed for signal processing,
which aims to find a shorter version of the signal and meanwhile  preserve the maximum information of the signal \cite{ib}. More recently, variational information bottleneck (VIB) \cite{vib}  bridges the gap between IB and deep learning, which approximates the information bottleneck constraints and enables IB to be applicable with deep learning. Nowadays, due to its ability to learn minimal informative representations, IB and VIB have been widely employed in computer vision \cite{ib-cv}, reinforcement learning \cite{ib-rl}, and natural language processing \cite{ib-nlp}.  Our paper is based on VIB, adapting and improving VIB for multimodal sentiment analysis and emotion recognition.

Among all the IB variants, the works that  aim to adopt IB for multi-view learning are the most related to our work \cite{mib-iclr,poe,MVAE,mvib,mvib2}.
Lee \textit{et al.} \cite{mvib}, Wu \textit{et al.} \cite{MVAE}, and Zhu \textit{et al.} \cite{mvib2} extend VIB for multi-view learning, which first constrain the information flow of each view using VIB, and then obtain the joint representation via the Product-of-Expert (PoE) \cite{poe}. Similarly, Wang \textit{et al.} develop a deep multi-view IB theory, aiming to maximize the mutual information between the labels and the learned joint representation and minimize the mutual information between the learned representation of each view and the original representation \cite{dmib}. The main differences between the proposed method and above multi-view IB methods are: \textbf{1)} instead of only applying the IB principle on each modality, we develop E-MIB, L-MIB and C-MIB to study multimodal information bottleneck more comprehensively. Our methods manage to reduce the redundancy of the generated multimodal representation via the E-MIB, and filter out noisy information of each modality via the L-MIB. Moreover, C-MIB manages to combine the advantages of L-MIB and E-MIB to regularize both unimodal and multimodal representations. In a word, the proposed method simultaneously applies the IB principle to unimodal representation learning and complicated multimodal representation learning, which enables the learning of more discriminative and expressive features; \textbf{2)} the proposed MIB variants are flexible as they can integrate any fusion mechanisms rather than only applying PoE as in previous multi-view IB methods \cite{poe,MVAE, mvib,mvib2}. In this way, more complicated and effective fusion strategy can be utilized to provide higher expressive power and explore more sufficient multimodal dynamics, which is shown to be important in our experiment (see Section~\ref{sec:fusion_e}). The compatibility of any fusion mechanisms is one of the advantages of the proposed method over multi-view IB methods.  Notably, Ma \textit{et al.} \cite{mib-iccv} propose a multimodal information bottleneck for image-to-speech synthesis. However, the multimodal information bottleneck in \cite{mib-iccv} is fundamentally different from our method, whose information bottleneck mainly refers to the designed model architecture instead of the IB principle.

\begin{figure*}
\setlength{\belowcaptionskip}{-0.3cm}
\centering
\includegraphics[scale=0.15]{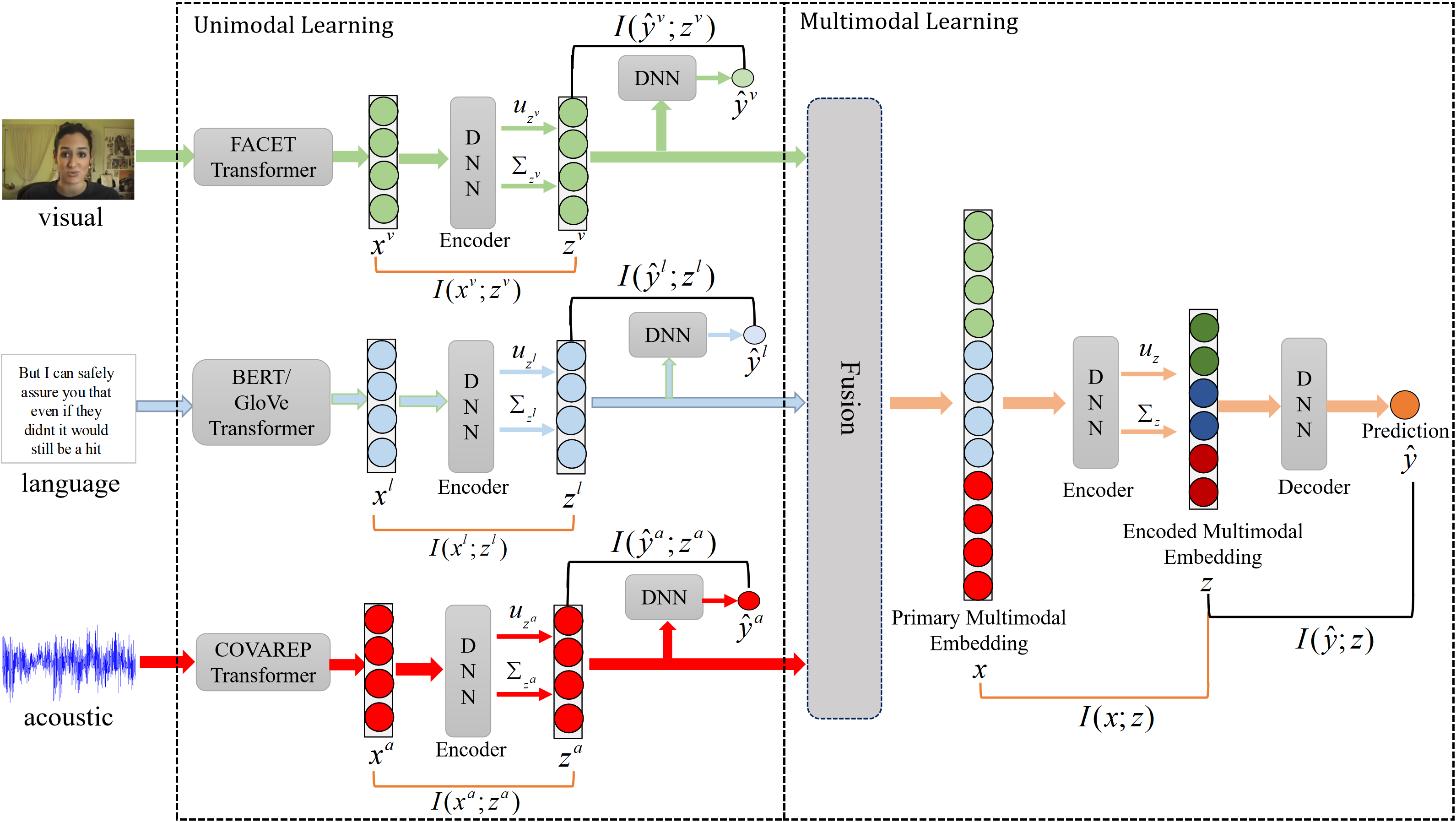}
\caption{\label{1}\textbf{The Diagram of C-MIB.} DNN denotes deep neural network. The fusion mechanism in our framework is flexible.
}
\vspace{-0.2cm}
\end{figure*}

\section{\textbf{Model Architecture}}\label{sec:Algorithm}
In this section, we introduce the pipeline of our MIB variants in detail, and the diagram of C-MIB is illustrated in Fig.~\ref{1}. MIB consists of three unimodal learning networks and a multimodal fusion network, where the optimization of these networks is driven by the information bottleneck principle. The pipelines of the three MIB variants are illustrated in Algorithm~\ref{alg1}, ~\ref{alg2}, and ~\ref{alg3}, respectively.
The detailed introduction is shown in the following sections.

\begin{algorithm}[h]
\caption{Early-fusion MIB (E-MIB)}
\raggedright
\label{alg1}
\textbf{input:} Unimodal representation $\bm{U}_m, \ m \in \{a, v, l \}$, hyper-parameter $\beta$\\
\textbf{output:} prediction $\hat{y}_i$\\
    \begin{algorithmic}[t]
		\State Initialize unimodal learning networks $F^m$ and multimodal fusion network $F^f$;
		\While {not done}
		     \State Sample a batch of utterances
		     \For{each utterance $i$}
		     \For{each of modality $m$\ \ ($m \in \{ l, a, v\})$}
		     \State $\bm{x}^{m}_i=\bm{F}^m(\bm{U}^m_i; \theta_m) $
		     \EndFor
		     \State $\bm{x}_i=\bm{F}^f(\bm{x}^{l}_i, \bm{x}^{a}_i, \bm{x}^{v}_i; \theta_f)$
		     \State $ \boldsymbol{\mu}_{z_i},  \boldsymbol{\Sigma}_{z_i} = \boldsymbol{\mu}(\bm{x}_i;\theta_\mu), \boldsymbol{\Sigma}(\bm{x}_i;\theta_\Sigma)$
		     \State $\bm{z}_i  = \boldsymbol{\mu}_{z_i}+ \boldsymbol{\Sigma}_{z_i}\times \boldsymbol{\epsilon}$
		     \State $\hat{y}_i = \bm{D}(\bm{z}_i; \theta_d)$
		     \EndFor	
		     \State Compute $J_{E-MIB}$ as in Eq.~\ref{E-MIB}
        \EndWhile
    \end{algorithmic}
\vspace{-0.1cm}
\end{algorithm}
\vspace{-0.1cm}

\begin{algorithm}[h]
\caption{Late-fusion MIB (L-MIB)}
\raggedright
\label{alg2}
\textbf{input:} Unimodal representation $\bm{U}_m, \ m \in \{a, v, l \}$, hyper-parameter $\beta$\\
\textbf{output:} prediction $\hat{y}_i$\\
\begin{algorithmic}[t]
		\State Initialize unimodal learning networks $F^m$ and multimodal fusion network $F^f$;
		\While {not done}
		     \State Sample a batch of utterances
		     \For{each utterance $i$}
		     \For{each of modality $m$\ \ ($m \in \{ l, a, v\})$}
		     \State $\bm{x}^{m}_i=\bm{F}^m(\bm{U}^m_i; \theta_m) $
		     \State $ \boldsymbol{\mu}_{z_i^m},  \boldsymbol{\Sigma}_{z_i^m} = \boldsymbol{\mu^m}(\bm{x}_i^m;\theta_\mu^m), \boldsymbol{\Sigma^m}(\bm{x}_i^m;\theta_\Sigma^m)$
		     \State $\bm{z}_i^m  = \boldsymbol{\mu}_{z_i^m}+ \boldsymbol{\Sigma}_{z_i^m}\times \boldsymbol{\epsilon}$	
		     \State $\hat{y}_i^m = \bm{D^m}(\bm{z}_i^m; \theta_d^m)$
		     \EndFor
		     \State $\bm{z}_i=\bm{F}^f(\bm{z}^{l}_i, \bm{z}^{a}_i, \bm{z}^{v}_i; \theta_f)$
		     \State $\hat{y}_i = \bm{D}(\bm{z}_i; \theta_d)$
		     \EndFor	
		     \State Compute $J_{L-MIB}$ as in Eq.~\ref{L-MIB}
		\EndWhile{}
\end{algorithmic}
\end{algorithm}
\vspace{-0.3cm}

\begin{algorithm}[h]
\caption{Complete MIB (C-MIB)}
\raggedright
\label{alg3}
\textbf{input:} Unimodal representation $\bm{U}_m, \ m \in \{a, v, l \}$, hyper-parameter $\beta$\\
\textbf{output:} prediction $\hat{y}_i$\\
\begin{algorithmic}[t]
		\State Initialize unimodal learning networks $F^m$ and multimodal fusion network $F^f$;
		\While {not done}
		     \State Sample a batch of utterances
		     \For{each utterance $i$}
		     \For{each of modality $m$\ \ ($m \in \{ l, a, v\})$}
		     \State $\bm{x}^{m}_i=\bm{F}^m(\bm{U}^m_i; \theta_m) $
		     \State $ \boldsymbol{\mu}_{z_i^m},  \boldsymbol{\Sigma}_{z_i^m} = \boldsymbol{\mu^m}(\bm{x}_i^m;\theta_\mu^m), \boldsymbol{\Sigma^m}(\bm{x}_i^m;\theta_\Sigma^m)$
		     \State $\bm{z}_i^m  = \boldsymbol{\mu}_{z_i^m}+ \boldsymbol{\Sigma}_{z_i^m}\times \boldsymbol{\epsilon}$	
		     \State $\hat{y}_i^m = \bm{D^m}(\bm{z}_i^m; \theta_d^m)$
		     \EndFor
		     \State $\bm{x}_i=\bm{F}^f(\bm{z}^{l}_i, \bm{z}^{a}_i, \bm{z}^{v}_i; \theta_f)$
		     \State $ \boldsymbol{\mu}_{z_i},  \boldsymbol{\Sigma}_{z_i} = \boldsymbol{\mu}(\bm{x}_i;\theta_\mu), \boldsymbol{\Sigma}(\bm{x}_i;\theta_\Sigma)$
		     \State $\bm{z}_i  = \boldsymbol{\mu}_{z_i}+ \boldsymbol{\Sigma}_{z_i}\times \boldsymbol{\epsilon}$
		     \EndFor	
		     \State Compute $J_{C-MIB}$ as in Eqs.~\ref{C-MIB1},~\ref{C-MIB2},~\ref{C-MIB3}
		\EndWhile{}
\end{algorithmic}
\end{algorithm}
\vspace{-0.3cm}

\subsection{\textbf{Notations and Task Definition}}

The downstream tasks in this paper are multimodal sentiment analysis and multimodal emotion recognition. The input to the model is an utterance \cite{Olson1977From}, i.e., a segment of a video bounded by a sentence. Each utterance has three modalities, i.e., acoustic ($a$), visual ($v$), and language ($l$) modalities. The input sequences of acoustic, visual, and language modalities are denoted as $\bm{U_a} \in \mathbb{R}^{T_a \times d_a}$, $\bm{U_v} \in \mathbb{R}^{T_v \times d_v}$, and $\bm{U_l} \in \mathbb{R}^{T_l \times d_l}$ respectively, where $T_m$ and $d_m$ denotes the sequence length and the feature dimensionality respectively ($m\in \{a, v, l \}$). Given the three unimodal sequences, we aim to learn a joint multimodal representation and then predict the sentiment score or emotion label based on the multimodal representation.

After being processed by unimodal learning network $F^m$, the input feature sequence $\bm{U_m} \in \mathbb{R}^{T_m \times d_m}$ becomes the unimodal representation $\bm{x_m} \in \mathbb{R}^{d_m}$. For conciseness, the structures of unimodal learning network $F^m$ and the fusion network $F^f$ are illustrated in the Appendix. Note that our algorithm is independent of the concrete fusion mechanism, and we can inject various fusion methods into the multimodal  fusion  network  to  provide  higher  expressive  power. In  this  paper,  we  mainly  investigate  five  fusion  methods  to verify the effectiveness of our algorithm and comprehensively  compare the performance between MIB variants.

\subsection{\textbf{Multimodal Information Bottleneck}}\label{sec:MIB}
In this section, we introduce the proposed MIB variants in detail. We extend the traditional information bottleneck (IB) to learn minimal sufficient multimodal representation, which contains the necessary information that is discriminative for the label and is free of noise. Moreover, we also add the IB constraint on the encoded unimodal representations to filter out the noisy information in unimodal representations that might interfere the learning of cross-modal interactions.

\subsubsection{\textbf{Information Bottleneck}}
Firstly, we introduce the general principle of information bottleneck (IB). IB aims to find better representation under the constraint on its complexity, such that the latent representation is discriminative and  simultaneously free of redundancy. The IB principle defines what we mean by a good representation, in terms of the fundamental tradeoff between the concise representation and the good predictive power \cite{vib}. Specifically, IB is based on Mutual information (MI), aiming to maximize the MI between the encoded representation and the label as well as minimize the MI between the encoded representation and the input.

MI measures the amount of information obtained in one random variable after observing another random variable. Formally, given two random variables $\bm{x}$ and $y$ with joint distribution $p(\bm{x},y)$ and marginal distributions $p(\bm{x})$ and $p(y)$, their MI is defined as the KL-divergence between the joint distribution and the product of their marginal distributions. The equation is shown as follows:
\begin{equation}
\label{eq88}
\begin{aligned}
I(\bm{x};y)&=I(y;\bm{x})=KL\Big(p(\bm{x},y)||p(\bm{x})p(y)\Big)\\
&=\int d\bm{x}dyp(\bm{x},y)\log\frac{p(\bm{x},y)}{p(\bm{x})p(y)}\\
&=\mathbb{E}_{(\bm{x},y)\sim p(\bm{x},y)}\bigg[\log\frac{p(\bm{x},y)}{p(\bm{x})p(y)}\bigg]
\end{aligned}
 \end{equation}

The goal of IB is to learn the compressed encoded representation $\bm{z}$ using the input $\bm{x}$, where $\bm{z}$ is maximally discriminative about the target $y$ (i.e., $I(y;\bm{z})$ is maximized). Obviously,
the maximally informative  representation can always be obtained by taking the identity mapping, i.e. $\bm{x}=\bm{z}$, which however contains noisy information or is often redundant for prediction. Therefore, a constraint is added on the MI between  $\bm{z}$ and $\bm{x}$. The objective of IB thereby becomes:
\begin{equation}
\begin{aligned}
  (1) \ \underset{}{\text{maximize }}  \   I(y;\bm{z})\ \ \ \ \text{and}\ \ \ \ (2)\ \underset{} {\text{minimize }}  \   I(\bm{x};\bm{z})
\end{aligned}
\end{equation}
where the first constraint encourages $\bm{z}$ to be maximally predictive of the target $y$, and the second constraint forces $\bm{z}$ to contain as less information from $\bm{x}$ as possible.
In other words, the IB principle aims to learn the minimal sufficient representation of $\bm{x}$ with respect to the label $y$, explicitly enforcing the $\bm{z}$ to only preserve the information in $\bm{x}$ that is discriminative to the prediction. In this paper, we refer the second constraint as the \textbf{minimal information constraint}. A common objective function for IB is defined as:
\begin{equation}
    L_{IB}=I(y;\bm{z})-\beta I(\bm{x};\bm{z})
\end{equation}
where $\beta\geq0$ is a scalar that determines the weight of the minimal information constraint during optimization.

\begin{figure}
\setlength{\belowcaptionskip}{-0.3cm}
\centering
\includegraphics[scale=0.11]{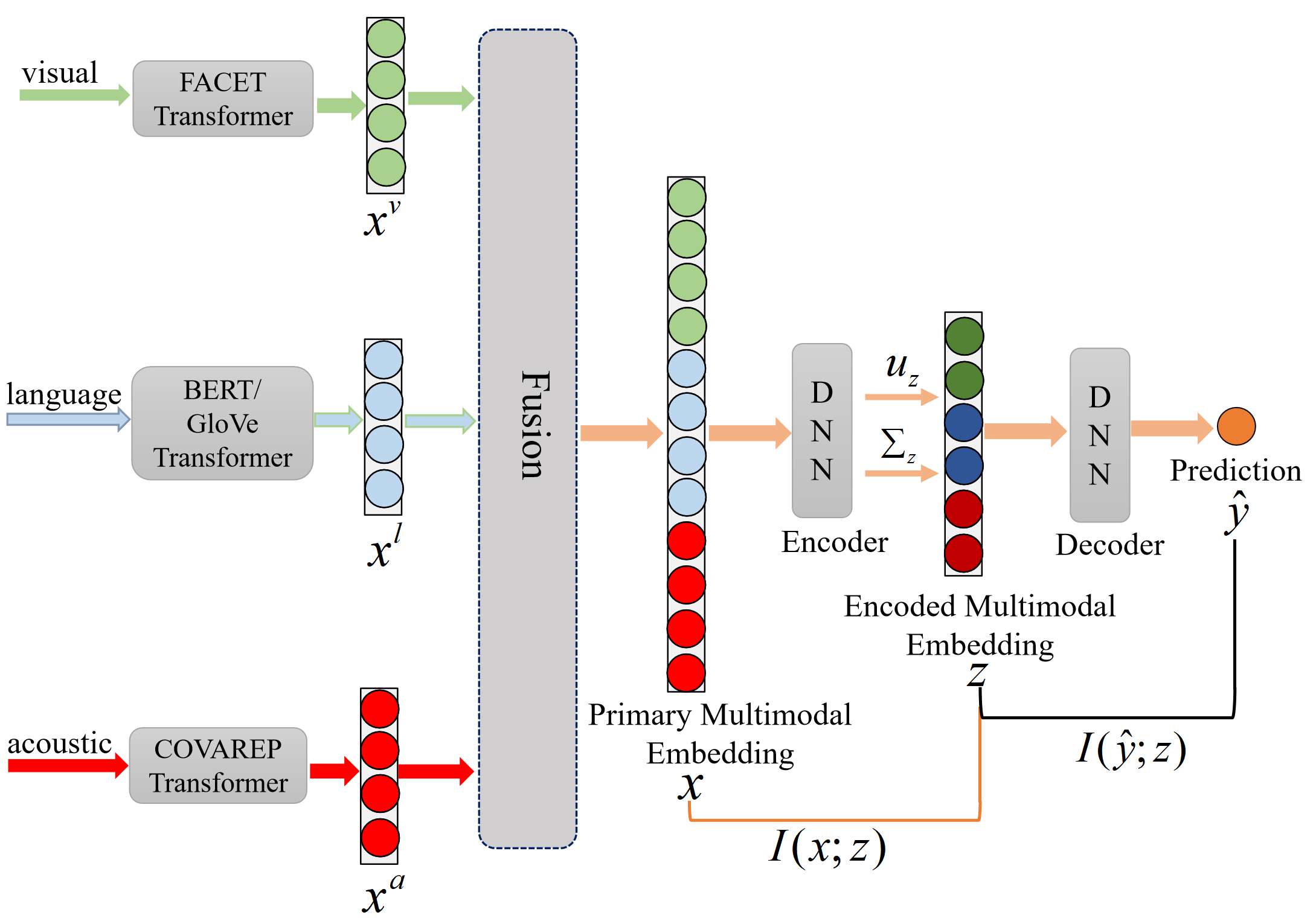}
\caption{\label{eemib}\textbf{The Schematic Diagram of E-MIB.}
}
\vspace{-0.3cm}
\end{figure}

\subsubsection{\textbf{Procedure of Early-Fusion MIB}}
Different from the traditional IB, in our multimodal system, the input contains three unimodal representations $\bm{x}^m$ output by the unimodal learning networks $\bm{F}^m$ ($m\in \{a,v,l \}$). How to enforce the learned  multimodal representation to contain as less information from the three unimodal representations as possible remains a problem. One intuitive way is to firstly fuse the unimodal representations into a joint multimodal representation, and then regularize the multimodal representation via IB, which is referred as early-fusion MIB (E-MIB). The pipeline of E-MIB is presented in Fig.~\ref{eemib}. In this case, the objective function of E-MIB can be defined as:
\begin{equation}
    L_{MIB}=I(y;\bm{z})-\beta I(\bm{x}^a, \bm{x}^v, \bm{x}^l;\bm{z})
\end{equation}
where $y$ is the label, $\bm{z}$ is the encoded multimodal representation, and $\bm{x}^a$, $\bm{x}^v$, $\bm{x}^l$ refer to the three unimodal representations output by the unimodal learning networks (see Appendix), respectively.
To resolve the problem of multiple input modalities, we firstly fuse the three modalities to obtain a primary multimodal representation $\bm{x}$ via the fusion mechanisms introduced in Appendix. Then we reduce the redundancy of the primary multimodal representation and filter out noisy information via the IB principle (that is why we call it early fusion). Specifically, the defaulted fusion mechanism
is defined as the Concatenation: $\bm{x} = \bm{x}^a \oplus \bm{x}^v \oplus \bm{x}^l$. By the Concatenation, the $\bm{x}$ can preserve all the information from the three modalities, which is  used to learn the encoded multimodal representation $\bm{z}$ that is minimal sufficient.
The objective function thus can be rewritten as:
\begin{equation}
\label{eqemibo}
\begin{aligned}
    L_{E-MIB}&=I(y;\bm{z})-\beta I(\bm{x}^a, \bm{x}^v, \bm{x}^l;\bm{z})\\
           & = I(y;\bm{z})-\beta I(\bm{x};\bm{z})\\
            \text{s.t.\ }&\bm{x} = F^f(\bm{x}^a, \bm{x}^v, \bm{x}^l;\ \theta_f)
\end{aligned}
\end{equation}

Note that our E-MIB differs from previous multi-view IB methods \cite{MVAE,mvib,mvib2} in two main perspectives: 1) previous methods tend to apply the IB principle on each view. In contrast, the proposed E-MIB can reduce the redundant information in the generated complex multimodal embedding. Actually, the multimodal embedding integrates more information and is often high-dimensional and redundant, which is directly used for prediction. Therefore, applying the IB principle on multimodal representation is more reasonable and effective; 2) existing multi-view IB methods \cite{MVAE,mvib,mvib2} mainly leverage PoE \cite{poe} to fuse information of each view to obtain the joint representation. Unlike them, our proposed E-MIB is flexible to integrate commonly-used fusion mechanisms in addition to PoE. In this way, more complicated and effective fusion strategies can be utilized to provide higher expressive power and explore more sufficient multimodal dynamics without redundancy. The compatibility of any commonly-used fusion mechanisms is one of the advantages of our MIB variants.

To optimize the objective function in Eq.~\ref{eqemibo}, we adopt the solution introduced in VIB \cite{vib}. For brevity, we place the deduction in Appendix and only show the result of deduction:
\begin{equation}
    I(y;\bm{z})
    \geq \int d\bm{x}dyd\bm{z} p(\bm{z}\mid \bm{x})p(y\mid \bm{x})p(\bm{x})\log q(y\mid \bm{z})
\end{equation}
where $q(y\mid \bm{z})$ is a variational approximation to $p(y\mid \bm{z})$. The entropy of the target label, i.e., $H(y)=- \int dy p(y)\log p(y)$ is independent of the parameter optimization and thereby can be ignored. By this means, we can instead turn to maximize the lower bound of the objective function to optimize $I(y;\bm{z})$.

For the second term of the E-MIB (i.e., the minimal information constraint), we also place the detailed deduction in the Appendix and only present the result of deduction:
\begin{equation}
\begin{aligned}
    I(\bm{x};\bm{z})
    \leq \int d\bm{x}d\bm{z} p(\bm{x})p(\bm{z}\mid \bm{x})\log \frac{p(\bm{z}\mid \bm{x})}{q(\bm{z})}
\end{aligned}
\end{equation}
where $q(\bm{z})$ is a variational approximation to the marginal distribution $p(\bm{z})$ which is often fixed to a standard normal Gaussian distribution.
Combining the above two constraints, we can obtain a lower bound of the objective function:
\begin{equation}
\label{eq1234}
\begin{aligned}
    &L_{E-MIB}=I(y;\bm{z})-\beta I(\bm{x};\bm{z})\geq J_{E-MIB}\\
    &\!=\! \mathbb{E}_{(\bm{x},y)\sim p(\bm{x},y), \bm{z}\sim p(\bm{z}\mid \bm{x})}\!\bigg[\!\log q(y\!\mid\! \bm{z})\!-\!\beta\!\cdot\! KL\Big(\!p(\bm{z}\!\mid\! \bm{x})||q(\bm{z})\!\Big)\!\bigg]
\end{aligned}
\end{equation}
where  $J_{E-MIB}$ is a lower bound of $L_{E-MIB}$. By maximizing $J_{E-MIB}$, the lower bound of $L_{E-MIB}$ is improved and thus $L_{E-MIB}$ can be optimized.

To use deep neural networks to optimize $J_{E-MIB}$, we assume that $p(\bm{z}\!\mid\! \bm{x})$ is a Gaussian distribution. The mean and variance of the Gaussian
distributions $p(\bm{z}\!\mid\! \bm{x})$ are learned using the respective deep neural network:
\begin{equation}
\label{eqpzx}
 p(\bm{z}\mid \bm{x})  = \mathcal{N}\bigg(\boldsymbol{\mu}(\bm{x};\theta_\mu), \boldsymbol{\Sigma}(\bm{x};\theta_\Sigma)\bigg) =
 \mathcal{N}\bigg(\boldsymbol{\mu}_z, \boldsymbol{\Sigma}_z\bigg)
\end{equation}
where $\boldsymbol{\mu}$ parameterized by $\theta_\mu$ and $\boldsymbol{\Sigma}$ parameterized by $\theta_\Sigma$ are the deep neural networks to learn the mean $\boldsymbol{\mu}_z$ and variance $\boldsymbol{\Sigma}_z$ of Gaussian distributions. However, in this way, the update of the parameters  will be a problem because the  computation of the gradients of parameters suffers from randomness. To resolve this problem, following \cite{dmib,vib}, we use the reparameterization trick to obtain $\bm{z}$:
\begin{equation}
 \bm{z}  = \boldsymbol{\mu}_z+ \boldsymbol{\Sigma}_z\times \boldsymbol{\epsilon}
\end{equation}
where $\boldsymbol{\epsilon} \sim \mathcal{N}(0,I)$ is a standard normal Gaussian distribution, and $I$ is the identity vector whose elements are all equal to 1. By using the reparameterization trick, the randomness is transferred to $\epsilon$ and the gradients of the parameters can be explicitly computed. Note that here we assume that each element in the vector $\bm{z}$ is independent from each other.

Moreover, for regression task, we formulate $q(y\mid \bm{z})$ as:
\begin{equation}
\begin{aligned}
  &q(y\mid \bm{z})  = e^{-||y-\boldsymbol{D}(\bm{z}; \theta_d)||_1+C}\\
  &\log q(y\mid \bm{z})  = -||y-\boldsymbol{D}(\bm{z}; \theta_d)||_1+C = -||y-\hat{y}||_1+C
\end{aligned}
\end{equation}
where $C$ is a constant, $\boldsymbol{D}$ is a decoder parameterized by $\theta_d$, and $\hat{y}$ is the prediction output by E-MIB. Under such circumstance, maximizing of $q(y\mid \bm{z})$ is equivalent to minimize the mean absolute error (MAE) between the prediction $\hat{y}$ and the target $y$\cite{mvib2}.
And for classification task, $\log q(y\mid \bm{z})$ is formulated as:
\begin{equation}
	\begin{aligned}
		& q(y\mid \bm{z}) = \hat{y}^y \cdot (1-\hat{y})^{1-y}, \ \ \hat{y} = \text{Sigmoid}(\boldsymbol{D}(\bm{z}; \theta_d)) \\
		&\log q(y\mid \bm{z})  = y\log \hat{y} + (1-y) log (1-\hat{y})
	\end{aligned}
\end{equation}
Here maximizing of $q(y\mid \bm{z})$ is equivalent to minimize the cross-entropy between the prediction $\hat{y}$ and the target $y$.
In general,  MAE and cross-entropy loss is used to maximize the mutual information between the target and the encoded multimodal representation $\bm{z}$ for regression task and classification task, respectively.
Furthermore, in practice, the approximated marginal distribution of the encoded multimodal representation $\bm{z}$ is often assumed to be a standard normal Gaussian distribution\cite{dmib,vib}:
\begin{equation}
\label{eqqz}
 q(\bm{z})  \sim \mathcal{N}(0,I)
\end{equation}

Combining Eq.~\ref{eqpzx} and Eq.~\ref{eqqz}, the KL-divergence term, i.e., $KL\Big(p(\bm{z}\!\mid\! \bm{x})||q(\bm{z})\Big)$ can be calculated as:
 \begin{equation}
\label{eq888}
\begin{aligned}
KL\Big(p(\bm{z}\!\mid\! \bm{x})||q(\bm{z})\Big)&\! =\!KL\Big(\mathcal{N}(\boldsymbol{\mu}(\bm{x};\theta_\mu), \boldsymbol{\Sigma}(\bm{x};\theta_\Sigma))||\mathcal{N}(0,I)\!\Big)\\
&=\! KL\Big(\mathcal{N}(\boldsymbol{\mu}_z, \boldsymbol{\Sigma}_z)||\mathcal{N}(0,I)\Big) \\
\end{aligned}
 \end{equation}
Here we assume the choice of reparameterization of $p(\bm{z}\!\mid \!\bm{x})$ and $q(\bm{z})$ allows for the computation of an analytic KL-divergence.  The KL-divergence of two Gaussian variables can be calculated by calculus and the computation is omitted here for conciseness.

Finally, the integral over $\bm{x}$, $\bm{z}$ and $y$ can be approximated by Monte Carlo sampling \cite{shapiro2003monte}, and thus $J_{E-MIB}$ (Eq.~\ref{eq1234}) can be simplified as:
\begin{equation}
	\small
	\label{E-MIB}
	\begin{aligned}
		&J_{E-MIB}\! \approx \! \frac{1}{n}\sum^{n}_{i=1}\Bigg[ \mathbb{E}_{\boldsymbol{\epsilon}\sim p(\boldsymbol{\epsilon})}[ \log q(y_i\!\mid\! \bm{z}_i)] -\! \beta\!\cdot \! KL\Big(p(\bm{z}_i\!\mid\! \bm{x}_i)|| q(\bm{z}_i)\Big)  \Bigg]\\
		&=\frac{1}{n}\sum^{n}_{i=1}\Bigg[ \mathbb{E}_{\boldsymbol{\epsilon}\sim p(\boldsymbol{\epsilon})}[ \log q(y_i\!\mid\! \bm{z}_i)]  -\! \beta\!\cdot KL\Big(\mathcal{N}(\boldsymbol{\mu}_{z_i}, \boldsymbol{\Sigma}_{z_i})||\mathcal{N}(0,I)\Big) \Bigg]
	\end{aligned}
\end{equation}
where $n$ is the sampling size (i.e., batch size), and $i$ is the subscript that indicates each sample (which is omitted in the above equations for conciseness). The concrete form of $\log q(y_i\!\mid\! \bm{z}_i)$ depends on the task type (regression or classification). By maximizing $J_{E-MIB}$, we encourage $\bm{z}$ to be maximally discriminative to the target and simultaneously forget the  information about the primary multimodal representation $\bm{x}$, by which means $\bm{z}$ ideally is no longer redundant.

\begin{figure}
\setlength{\belowcaptionskip}{-0.3cm}
\centering
\includegraphics[scale=0.09]{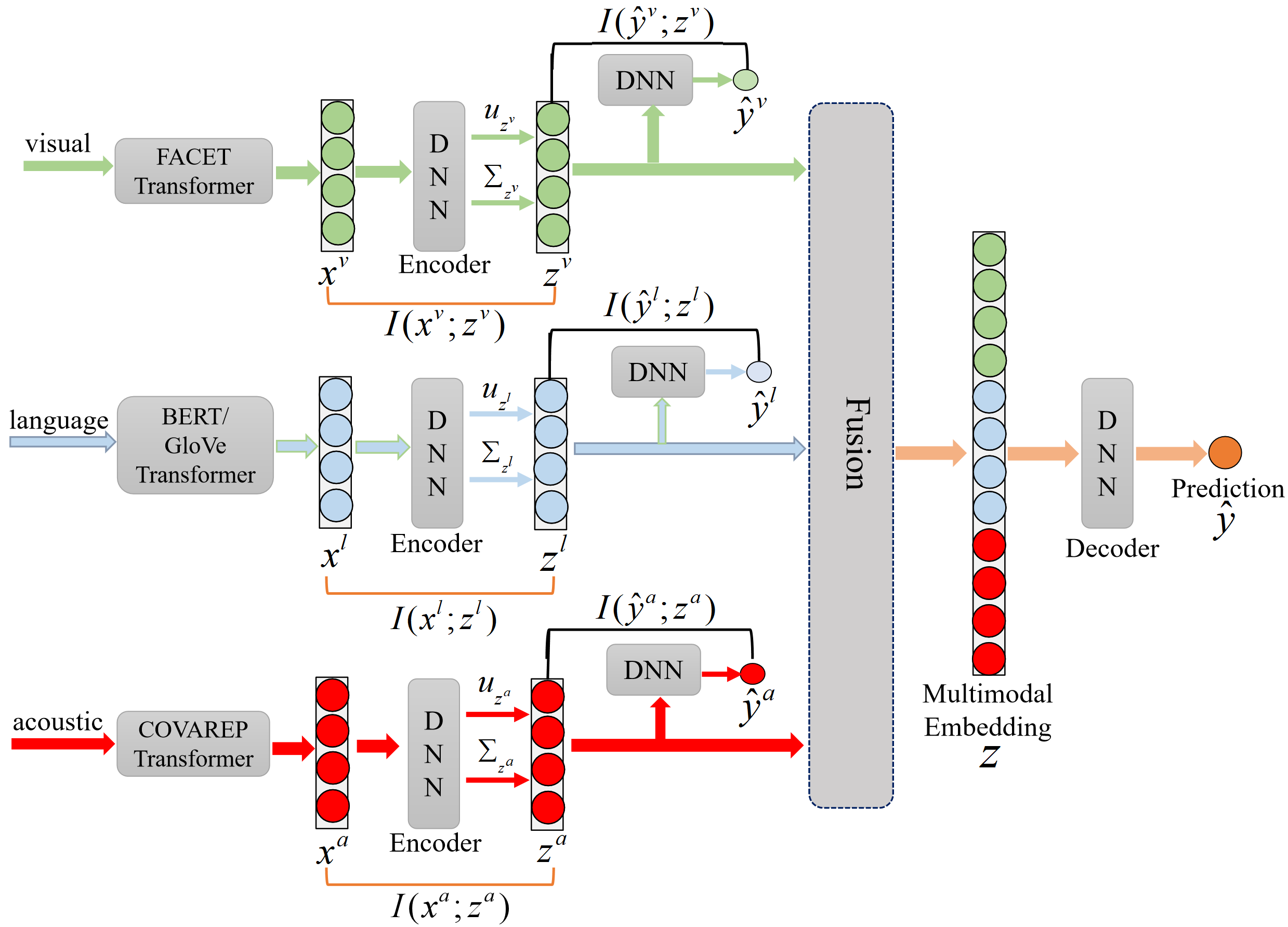}
\caption{\label{llmib}\textbf{The Schematic Diagram of L-MIB.}
}
\vspace{-0.3cm}
\end{figure}

\subsubsection{\textbf{Procedure of Late-Fusion MIB (L-MIB)}}

A disadvantage of E-MIB is that unimodal features might contain noise that negatively affects the exploration of cross-modal interactions during fusion. Moreover, the distribution discrepancy of unimodal representations might seriously prevent the model from discovering complementary information between modalities. Therefore, we propose another version of MIB, namely late-fusion MIB (L-MIB), which allows  unimodal representations to filter out noisy information before fusion and match the encoded unimodal distributions by individually applying the IB principle on each modality. The pipeline of L-MIB is shown in Fig.~\ref{llmib}, and the objective of L-MIB is:
\begin{equation}
\begin{aligned}
    L_{L-MIB} =& I(y; \bm{z})+ \sum_m [I(y;\bm{z}^m)- \beta I(\bm{x}^m;\bm{z}^m)]\\
    & \text{s.t.\ } \bm{z} = F^f(\bm{z}^a, \bm{z}^v, \bm{z}^l; \theta_f)
\end{aligned}
\end{equation}
where $F^f$ is the multimodal fusion network described in the Appendix.
From the deduction above, $L_{L-MIB}$ can be approximated by the following objective:
\begin{equation}
	\small
	\label{L-MIB}
	\begin{aligned}
		J_{L-MIB}& \approx\frac{1}{n}\sum^{n}_{i=1}\Bigg[\log q(y_i\mid \bm{z}_i) +\\ & \sum_m\! \Big[\! \mathbb{E}_{\boldsymbol{\epsilon}\sim p(\boldsymbol{\epsilon})}[ \log q(y_i\!\mid\! \bm{z}_i^m)] -\! \beta\!\cdot\! KL\Big(p(\bm{z}_i^m\!\mid\! \bm{x}_i^m)|| q(\bm{z}_i^m)\Big)\! \Big]\! \Bigg]\\
		&=\frac{1}{n}\sum^{n}_{i=1}\Bigg[ \log q(y_i\mid \bm{z}_i) +
		\sum_m \Big[ \mathbb{E}_{\boldsymbol{\epsilon}\sim p(\boldsymbol{\epsilon})}[ \log q(y_i\!\mid\! \bm{z}_i^m)] \\ & -\beta\cdot
		KL\Big(\mathcal{N}(\boldsymbol{\mu}_{z_i^m}, \boldsymbol{\Sigma}_{z_i^m})||\mathcal{N}(0,I)\Big)
		\Big] \Bigg]
	\end{aligned}
\end{equation}
where,
\begin{equation}
\begin{aligned}
 p(\bm{z}^m\mid \bm{x}^m)&  = \mathcal{N}\big(\boldsymbol{\mu^m}(\bm{x}^m;\theta_\mu^m), \boldsymbol{\Sigma^m}(\bm{x^m};\theta_\Sigma^m)\big)\\ & =
 \mathcal{N}\big(\boldsymbol{\mu}_{z^m}, \boldsymbol{\Sigma}_{z^m}\big)\\
  \bm{z}^m  &= \boldsymbol{\mu}_{z^m}+ \boldsymbol{\Sigma}_{z^m}\times \boldsymbol{\epsilon}
\end{aligned}
\end{equation}
where $\boldsymbol{D^m}$ is the decoder for modality $m$, $\boldsymbol{\mu^m}$ and $\boldsymbol{\Sigma^m}$ are the deep neural networks that learn the mean and the variance of $p(\bm{z}^m\mid \bm{x}^m)$ respectively.
From the perspective of filtering out noisy information before fusion, L-MIB is effective as it forces the encoded unimodal representation $\bm{z}^m$ to only contain the relevant information to the target. Moreover, by using the reparameterization trick, the distributions of different encoded unimodal representations are closer, which narrows down the modality gap.  The encoded unimodal representations are then fused to obtain the final multimodal representation $\bm{z}$, where $\bm{z}$ is used to infer the final prediction.

Similarly, the fusion procedure is flexible and we can integrate any fusion methods into L-MIB to obtain the encoded multimodal representation $\bm{z}$. Actually, another way to obtain $\bm{z}$  for L-MIB is referred to as the Bayesian product-of-experts (PoE) system \cite{poe,dmib,MVAE}:
\begin{equation}
	\begin{aligned}
		\boldsymbol{\mu}_{z} =& (\sum_m \boldsymbol{\mu}_{z^m} \cdot \boldsymbol{\Sigma}_{z^m}^{-1})(I +\sum_m \boldsymbol{\Sigma}_{z^m}^{-1})^{-1}\\
		&\boldsymbol{\Sigma}_{z} = (I +\sum_m \boldsymbol{\Sigma}_{z^m}^{-1})^{-1} \\
		&\bm{z}  = \boldsymbol{\mu}_{z}+ \boldsymbol{\Sigma}_{z}\times \boldsymbol{\epsilon}
	\end{aligned}
\end{equation}
where $\cdot$ denotes element-wise multiplication. PoE is similar to weighted average. The
mean part $\boldsymbol{\mu}_{z}$ of the  $\bm{z}$
is actually the average of modality-specific $\boldsymbol{\mu}_{z^m}$ weighted
by the reciprocal of the corresponding variance $\boldsymbol{\Sigma}_{z^m}$. PoE assumes the experts with greater confidence are the modalities whose information bears less uncertainty and is more relevant to the target. For experts with greater confidence (i.e., lower variance),  they will have larger weights to obtain $\bm{z}$ than those experts with higher variance. PoE severs as a fusion baseline and we will compare it with other fusion methods in Section~\ref{sec:fusion_e}.

Incorporating the IB principle for each modality helps to filter out the noise and focus on the discrimiative unimodal features. Especially for the acoustic and visual  modalities, they contain much noisy information that interferes the prediction\cite{TASLP}, which will negatively affect the quality of the fused multimodal representation (especially when complex fusion methods are applied). Therefore, it is beneficial to filter out the noise in the representations of these modalities.  However, by applying the IB principle on unimodal branches, the concurrence of the information or the complementary information between multiple modalities (which is not necessarily discriminative) might be ignored. Nevertheless, we argue that the  hyperparameter $\beta$ in  the proposed framework can alleviate this issue to some extent, balancing the non-discriminative  and discriminative information existing in different modalities.

\subsubsection{\textbf{Procedure of C-MIB}}
C-MIB is built to combine the advantages of E-MIB and L-MIB. Firstly, C-MIB applies the IB principle to extract encoded unimodal representations that ideally are free of noise. Then the encoded unimodal representations are fused to obtain the primary multimodal representation $\bm{x}$. The IB principle is further applied on $\bm{x}$ to reduce the redundancy and obtain the final multimodal representation $\bm{z}$. The objective of C-MIB is shown in:
\begin{equation}
\label{C-MIB_t}
\begin{aligned}
    L_{C-MIB} =& I(y; \bm{z}) - \beta I(\bm{x};\bm{z}) +\\ &\sum_m [I(y;\bm{z}^m)- \beta I(\bm{x}^m;\bm{z}^m)]\\
     \text{s.t.\ } \bm{x} =& F^f(\bm{z}^a, \bm{z}^v, \bm{z}^l; \theta_M)
\end{aligned}
\end{equation}
From the deduction above, $L_{C-MIB}$ can be approximated using $J_{C-MIB}$:
\begin{equation}
\label{C-MIB1}
\begin{aligned}
J_{C-MIB} \approx\frac{1}{n}\sum^{n}_{i=1}\Bigg[ J^M_i + \sum_m  J^m_i  \Bigg]
\end{aligned}
 \end{equation}
where,
\begin{equation}
	\label{C-MIB2}
	\begin{aligned}
		&J^{M}_i=\mathbb{E}_{\boldsymbol{\epsilon}\sim p(\boldsymbol{\epsilon})}[ \log q(y_i\!\mid\! \bm{z}_i)]  - \beta\cdot KL\Big(p(\bm{z}_i\mid \bm{x}_i)|| q(\bm{z}_i)\Big) \\
		&= \mathbb{E}_{\boldsymbol{\epsilon}\sim p(\boldsymbol{\epsilon})}[ \log q(y_i\!\mid\! \bm{z}_i)]  - \beta\cdot KL\Big(\mathcal{N}(\boldsymbol{\mu}_{z_i}, \boldsymbol{\Sigma}_{z_i})||\mathcal{N}(0,I)\Big)
	\end{aligned}
\end{equation}

\begin{equation}
	\label{C-MIB3}
	\begin{aligned}
		&J^{m}_i= \mathbb{E}_{\boldsymbol{\epsilon}\sim p(\boldsymbol{\epsilon})}[ \log q(y_i\!\mid\! \bm{z}_i^m)]  - \beta\cdot KL\Big(p(\bm{z}_i^m\mid \bm{x}_i^m)|| q(\bm{z}_i^m)\Big) \\
		&=\mathbb{E}_{\boldsymbol{\epsilon}\sim p(\boldsymbol{\epsilon})}[ \log q(y_i\!\mid\! \bm{z}_i^m)]  \!-\!\beta\cdot KL\Big(\mathcal{N}(\boldsymbol{\mu}_{z_i^m}, \boldsymbol{\Sigma}_{z_i^m})||\mathcal{N}(0,I)\Big)
	\end{aligned}
\end{equation}
The comparison of E-MIB, L-MIB and C-MIB is shown in experiment section.

\section{\textbf{Experiments}}\label{sec:Experiments}
In this section, we mainly evaluate our model on the multimodal sentiment analysis task. To evaluate the generalization ability of our model to other multimodal learning tasks, we additionally present the results of our model on the multimodal emotion recognition task.

\subsection{Datasets}

\subsubsection{\textbf{CMU-MOSI}}
CMU-MOSI \cite{Zadeh2016Multimodal} is a widely-used dataset for multimodal sentiment analysis. It contains 93 videos in total, and each video is divided into 62 utterances at most. The intensity of sentiment of each utterance is within [-3,3], where -3 indicates the strongest negative sentiment, and +3 the strongest positive.

\subsubsection{\textbf{CMU-MOSEI}}
CMU-MOSEI \cite{MOSEI} is a large multimodal sentiment analysis and emotion recognition dataset that contains a total number of 2928 videos. The dataset has been segmented at the utterance level, and each utterance has been scored on two levels: sentiment ranging between [-3, 3], and emotion with six different values (i.e., anger, disgust, fear, happy, sad, and surprise). We predict the sentiment and emotion score of the utterance for this dataset.

\subsubsection{\textbf{IEMOCAP}}
IEMOCAP \cite{Busso2008IEMOCAP} is a multimodal emotion recognition dataset that contains a total number of 151 videos from 10 speakers. The videos are segmented into about 10K utterances. The dataset has the following labels: anger, happiness, sadness, neutral, excitement, frustration, fear, surprise and other. We take the first four emotions to compare with our baselines. We follow previous works \cite{TASLP,MULT} to report the classification accuracy and the F1 score of each emotion.

\subsection{Experimental Details}

We develop our model using PyTorch on GTX2080Ti with CUDA 10.1 and torch 1.4.0. The defaulted fusion method is set to Concatenation for all the MIB variants.
More experimental details, evaluation protocol, and the introduction of baselines are shown in the Appendix for lack of space.

\begin{table}[t]
\centering
 \caption{ \label{t1}\textbf{ Comparison with baselines on CMU-MOSI.}  `B-MIB' refers to the baseline MIB model whose objective is maximizing the mutual information between multimodal representation and the label, which is implemented by removing the minimal information constraint from E-MIB.
 }
\resizebox{.99\columnwidth}{!}{\begin{tabular}{c|c|c|c|c|c}
 \hline
    & Acc7 ($\uparrow$) & Acc2 ($\uparrow$) & F1 ($\uparrow$) & MAE  ($\downarrow$) & Corr ($\uparrow$) \\
 \hline
 EF-LSTM  & 31.6 & 75.8 & 75.6 & 1.053 & 0.613  \\
 LF-LSTM  & 31.6 & 76.4 & 75.4 & 1.037 & 0.620  \\
TFN \cite{Zadeh2017Tensor} & 32.2 & 76.4  & 76.3 & 1.017 & 0.604 \\
 LMF \cite{Liu2018Efficient} & 30.6 & 73.8  & 73.7 & 1.026 & 0.602 \\
 MFN \cite{Zadeh2018Memory} & 32.1 & 78.0  & 76.0 & 1.010 &0.635  \\
RAVEN \cite{RAVEN} & 33.8 & 78.8  & 76.9 & 0.968 & 0.667 \\
 MULT \cite{MULT} & 33.6 & 79.3  & 78.3 & 1.009 & 0.667 \\
 QMF \cite{Quantum} & 35.5 & 79.7  & 79.6 & 0.915 & 0.696 \\
 MISA \cite{MISA} & 42.3 &  83.4 & 83.6 & 0.783 & 0.761 \\
 MAG-BERT \cite{MAG-BERT} & 42.9 & 83.5 & 83.5 & 0.790 & 0.769 \\
 TFR-Net \cite{tfr-net} & 42.6 & 84.0 & 83.9 & 0.787 & 0.788 \\
 \hline
B-MIB & 46.6 & 82.8 & 82.8 & 0.737 &  0.788 \\
E-MIB  & \textbf{48.6 } & \textbf{85.3}  & \textbf{85.3} & \textbf{0.711} & \textbf{0.798} \\
L-MIB  & 45.8 & 84.6  & 84.6 & 0.732 & 0.790 \\
C-MIB  & $\underline{48.2}$  & $\underline{85.2}$  &  $\underline{85.2}$ &  $\underline{0.728}$ & $\underline{0.793}$  \\
 \hline
 \end{tabular}}
 \vspace{-0.1cm}
\end{table}%

\begin{table}[t]
\centering
 \caption{ \label{t2}\textbf{ Comparison with baselines on CMU-MOSEI.}
 }
\resizebox{.99\columnwidth}{!}{\begin{tabular}{c|c|c|c|c|c}
 \hline
        & Acc7 ($\uparrow$) & Acc2 ($\uparrow$) & F1 ($\uparrow$) & MAE  ($\downarrow$) & Corr ($\uparrow$) \\
 \hline
 EF-LSTM  & 46.7 & 79.1 & 78.8 & 0.665 & 0.621\\
 LF-LSTM  & 49.1 & 79.4 & 80.0 & 0.625 & 0.655 \\
 TFN \cite{Zadeh2017Tensor} & 49.8 & 79.4  & 79.7 & 0.610 & 0.671 \\
 LMF \cite{Liu2018Efficient} & 50.0 & 80.6  & 81.0 & 0.608 & 0.677 \\
 MFN \cite{Zadeh2018Memory} & 49.1 & 79.6  & 80.6 & 0.618 &0.670 \\
 RAVEN \cite{RAVEN} & 50.2 & 79.0  & 79.4 & 0.605 & 0.680 \\
 MULT \cite{MULT} & 48.2 & 80.2  & 80.5 & 0.638 & 0.659 \\
 IMR \cite{MRM} & 48.7 & 80.6  & 81.0 & - & - \\
 QMF \cite{Quantum} & 47.9 & 80.7  & 79.8 & 0.640 & 0.658 \\
  MISA \cite{MISA} & 52.2  &  85.5 & 85.3 & \textbf{0.555} & 0.756 \\
 MAG-BERT \cite{MAG-BERT} & 51.9 & 85.0 & 85.0 & 0.602 & 0.778 \\
  TFR-Net \cite{tfr-net} & 51.7 & 85.2 & 85.1 & 0.606 & 0.781 \\
 \hline
B-MIB & 52.1 & 84.4 & 84.3 & 0.603 & 0.775\\
E-MIB  & $\underline{53.2}$ & 85.4  & 85.4 & 0.588 & \textbf{0.790} \\
L-MIB  & \textbf{54.1} & $\underline{85.7}$  & $\underline{85.7}$ & 0.600 & 0.788 \\
C-MIB  & 53.0 & \textbf{86.2}  & \textbf{86.2} &  $\underline{0.584}$ & $\underline{0.789}$ \\
 \hline
 \end{tabular}}
 \vspace{-0.1cm}
\end{table}%

\subsection{\textbf{Comparison on Multimodal Sentiment Analysis}}
In this section, we compare our proposed MIB algorithms with our baselines on  the multimodal sentiment analysis task. The results are shown in Table~\ref{t1} and ~\ref{t2}, where the best results are highlighted in bold and the second best results are marked with underlines. For CMU-MOSI dataset, although TFR-Net\cite{tfr-net} performs better than the existing methods and sets up a high baseline, our MIB variants still obtain the best performance on the majority of the evaluation metrics. Specifically, E-MIB achieves the best results on all metrics, and it outperforms TFR-Net\cite{tfr-net} by 6.0\% on Acc7, 1.3\% on Acc2 and 1.4\% on F1 score. On CMU-MOSEI dataset, our proposed C-MIB yields 0.7\% improvement on Acc2 and 1.2\% improvement on F1 score compared to the current state-of-the-art  MISA \cite{MISA}. The performance on Acc7 and Corr is also improved compared to the state-of-the-art methods.
These results demonstrate the effectiveness of our MIB variants, indicating the importance of learning minimal sufficient representations.

Notably, to compare with baselines, we only present the results of our MIB variants with the fusion method being Concatenation. Interestingly, combining the performance on the two datasets, E-MIB performs comparably to C-MIB. This is reasonable because when Concatenation is applied, the unimodal dynamics cannot fully interact with each other, and thus the noise in unimodal representations will not seriously influence the quality of the multimodal representation. Therefore, there is no significant gain when the IB principle is applied on the unimodal representations. Actually, although our model is agnostic of fusion architectures, the specific fusion method can still differently influence the performance of different MIB variants. Therefore, we propose to compare the performance between the MIB variants by comprehensively estimating their performance  under different fusion strategies. For the comprehensive comparison between the MIB variants, please refer to Section~\ref{sec:fusion_e}. Also notice that the performance difference between the two different datasets exists. For instance, L-MIB performs weaker than E-MIB on CMU-MOSI, but it manages to perform better than E-MIB on CMU-MOSEI. The performance difference may be partly due to the reason that our methods introduce randomness when generating the latent representations, which might affect the performance, especially on a smaller dataset (i.e. CMU-MOSI). We also refer the reader to Section~\ref{sec:fusion_e} for the comprehensive comparison between the MIB variants, where the average performances of the MIB variants become consistent when various fusion methods are considered.

\begin{table}[t]
\small
\centering
 \caption{ \label{t544}\textbf{ Comparison with baselines on IEMOCAP.} }
\resizebox{1.0\columnwidth}{!}{\begin{tabular}{c|c|c|c|c|c|c|c|c}
 \hline
  &\multicolumn{2}{c|}{Happy} & \multicolumn{2}{c|}{Sad} & \multicolumn{2}{c|}{Angry} & \multicolumn{2}{c}{Neutral}\\
 \hline
  Models & Acc  & F1  & Acc  & F1  & Acc  & F1  & Acc  & F1 \\
 \hline
 MFN \cite{Zadeh2018Memory}& 86.5 & 84.0 & 83.5 & 82.1 & 85.0 & 83.7 & 69.6 & 69.2 \\
 Graph-MFN \cite{MOSEI}& 86.8 & 84.2 & 83.8 & 83.0 & 85.8 & 85.5 & 69.4 & 68.9 \\
 RAVEN \cite{RAVEN}& 87.3 & 85.8 & 83.4 & 83.1 & 87.3 & 86.7 & 69.7 & 69.3 \\
 LMF \cite{Liu2018Efficient}& 86.9 & 82.3 & 85.4 & 84.7 & 87.1 & 86.8 & \underline{71.6} & \underline{71.4} \\
 MULT \cite{MULT}& 87.4 & 84.1 & 84.2 & 83.1 & 88.0 & 87.5 & 69.9 & 68.4 \\
 HFFN \cite{HFFN}& 86.8 & 82.1 & 84.4 & 84.5 & 86.6 & 85.8 & 69.6 & 69.3 \\
  TCM-LSTM \cite{TASLP} & 87.2 & 84.8 & 84.4 & 84.9 & \underline{89.0} & 88.6 & 71.3 & 71.2 \\
  \hline
  E-MIB &  \underline{88.1}& \underline{85.9} & 86.8 & \textbf{86.9}  & 88.8 & \underline{88.7} & 71.0 & 70.9 \\
   L-MIB & \textbf{88.4} & \textbf{86.5} & \underline{86.9} & \underline{86.1} & 88.3 & 88.3 & 70.9 & 70.6 \\
   C-MIB & 87.7 & 85.8 & \textbf{87.6} & \textbf{86.9} & \textbf{89.2} & \textbf{88.8} & \textbf{72.4} & \textbf{72.1} \\
 \hline
 \end{tabular}}
 \vspace{-0.3cm}
\end{table}%

\begin{table*}[t]
\centering
\caption{ \label{t545}\textbf{ Comparison with baselines on CMU-MOSEI dataset (multimodal emotion recognition task).} }
\resizebox{1.9\columnwidth}{!}{\begin{tabular}{c|c|c|c|c|c|c|c|c|c|c|c|c|c|c}
 \hline
  &\multicolumn{2}{c|}{Anger} & \multicolumn{2}{c|}{Disgust} & \multicolumn{2}{c|}{Fear} & \multicolumn{2}{c}{Happy}& \multicolumn{2}{c|}{Sad} & \multicolumn{2}{c|}{Surprise} & \multicolumn{2}{c}{Average}\\
 \hline
  Models & W-Acc & AUC & W-Acc & AUC& W-Acc & AUC& W-Acc & AUC& W-Acc & AUC& W-Acc & AUC& W-Acc & AUC\\
 \hline
 EF-LSTM & 58.5 & 62.2 & 59.9 & 63.9 & 50.1  & 69.8  & 65.1  &68.9 & 55.1 & 58.6 &  50.6 & 54.3  & 56.6  & 63.0\\
 LF-LSTM & 57.7 & 66.5 & 61.0 & 71.9 & 50.7  & 61.1  &  63.9 & 68.6& 54.3 & 59.6 & 51.4  &  61.5 &  56.5 & 64.9\\
 Graph-MFN \cite{MOSEI} & 62.6 &  -& 69.1 & - & 62.0  & -  &\underline{66.3}   &- &  60.4 & - & 53.7  & -  &  62.3 & -\\
   MTL \cite{MTL} & 66.8 & 68.0 & 72.7 & 76.7 & 62.2  & 42.9  &  53.6 & 71.4 & 61.4 & 57.6 & 60.6  & 65.1  & 62.8  & 63.6\\
  MTEE \cite{MTE} & \underline{67.0} & \underline{71.7} & 72.5 & 78.3 & 65.4  & \textbf{71.6}  &  \textbf{67.9} & 73.9& \underline{62.6} & 66.7 &   62.1 & 66.4  & 66.2  & 71.2\\
  \hline
   E-MIB & \textbf{68.4} & 71.6 & \textbf{77.5} & \underline{79.0} & \textbf{70.6} & 69.1 & 55.3 & \underline{74.7} & 61.8 & \underline{67.9} &  \underline{66.3} & \textbf{68.0}  & \textbf{66.7}  & \underline{71.7}\\
  L-MIB & 66.9 & 71.5 & \underline{76.5} & 78.2 & \underline{69.5}  &  71.3 & 55.2  &  74.6 & \textbf{64.4} & 66.7 &  \textbf{66.7} &  66.4 &  \underline{66.5} & 71.4 \\
   C-MIB & 65.2  & \textbf{72.5}  & 75.3  & \textbf{79.8}  &  68.5 & \underline{71.5} & 55.2 & \textbf{75.2} & 61.7 & \textbf{68.7} &  65.4 & \underline{67.7}  & 65.9  & \textbf{72.5}\\
 \hline
 \end{tabular}}
  \vspace{-0.3cm}
\end{table*}%

\subsection{\textbf{Results on Multimodal Emotion Recognition}}
We additionally evaluate the MIB variants on the task of multimodal emotion recognition to justify the generalization ability of the models to other multimodal task. Two widely-used datasets, i.e. CMU-MOSEI and IEMOCAP are evaluated in this section.
For IEMOCAP,
as shown in Table~\ref{t544}, our C-MIB achieves the best performance in the task of recognizing the `Angry', `Sad', and `Neutral' emotions. Moreover, the L-MIB reaches the best performance of recognizing the `Happy' emotion. The E-MIB achieves the best result on the F1 score of recognizing the `Sad' emotion, yielding 2.0\% improvement compared with the best result of existing methods. Notably, combining the performance on all the emotions, all the MIB variants outperform the baselines.
For CMU-MOSEI, we can infer from Table~\ref{t545} that all MIB variants obtain very competitive performance on the average weighted acc and auc metrics. Our methods also outperform existing methods on the recognition of the majority of emotions.
The extra experiments of multimodal emotion recognition have proven the effectiveness and generalization ability of our methods.

\begin{table}[t]
\centering
 \caption{ \label{t3}\textbf{ Ablation studies on the CMU-MOSI dataset.} }
\resizebox{.95\columnwidth}{!}{\begin{tabular}{c|c|c|c|c|c}
 \hline
 & Acc7 ($\uparrow$) & Acc2 ($\uparrow$) & F1 ($\uparrow$) & MAE  ($\downarrow$) & Corr ($\uparrow$) \\
 \hline
 W/O IB on Language & $\underline{47.4} $&$ \underline{84.4}$ & $\underline{84.4}$ & \textbf{0.725} &  \textbf{0.794} \\
 W/O IB on Acoustic & 47.2 & 84.0 & 84.0 & \textbf{0.725} & 0.788  \\
 W/O IB on Visual & 46.4 & 84.0 & 84.0 & 0.732 & 0.791  \\
 \hline
C-MIB  & \textbf{48.2}  & \textbf{85.2}  &  \textbf{85.2} &  $\underline{0.728}$ & $\underline{0.793}$  \\
  \hline
 \end{tabular}}
 \vspace{-0.3cm}
\end{table}%

\subsection{\textbf{Ablation Study}}

In this section, we perform ablation studies on the C-MIB to investigate the contribution of each unimodal minimal information constraint loss by removing it from each modality, and the results are presented in Table~\ref{t3}. We design three ablation experiments to investigate the contribution of each component in our proposed C-MIB. Firstly, we remove the IB principle from language modality (see the case of `W/O IB on Language' where the language representation is directly used to generate multimodal representation $\bm{x}$), and the performance of the model has a slight drop. Secondly, we remove the IB principle on acoustic and visual modality respectively, and significant drop on performance is observed. These results demonstrate that the IB principle is more effective  on visual and acoustic modalities, mainly for the reason that they have many redundant and noisy information that is irrelevant to the prediction. Since the language modality is far more discriminative \cite{TASLP,MCTN,HFFN}, the influence is not that significant when the IB principle is removed from the language modality.

\begin{table*}[t]
\centering
 \caption{ \label{t4}\textbf{ Discussion on the fusion strategies and the MIB variants on CMU-MOSI dataset.} The best average results across all the fusion methods are highlighted, and the second best average results are marked with underlines.
 }
\resizebox{1.95\columnwidth}{!}{\begin{tabular}{c|c|c|c|c|c|c|c|c|c|c|c|c|c|c|c}
 \hline
     &  \multicolumn{5}{c|}{E-MIB}   &   \multicolumn{5}{c|}{L-MIB}  &  \multicolumn{5}{c}{C-MIB}  \\
 \hline
     & Acc7 ($\uparrow$) & Acc2 ($\uparrow$) & F1 ($\uparrow$) & MAE  ($\downarrow$) & Corr ($\uparrow$)   & Acc7 ($\uparrow$) & Acc2 ($\uparrow$) & F1 ($\uparrow$) & MAE  ($\downarrow$) & Corr ($\uparrow$)   & Acc7 ($\uparrow$) & Acc2 ($\uparrow$) & F1 ($\uparrow$) & MAE  ($\downarrow$) & Corr ($\uparrow$) \\
 \hline
 Concatenation &  48.6 & 85.3 & 85.3 & 0.711 & 0.798 & 45.8  & 84.6 & 84.6 & 0.732 & 0.790 & 48.2 & 85.2 & 85.2 & 0.728 &  0.793 \\
 Addition  & 47.3 & 84.3  & 84.3 & 0.714 & 0.799 & 45.6 & 84.4  & 84.4 & 0.722 & 0.793 & 48.2 &  84.6 & 84.6  & 0.721 & 0.796 \\
  Multiplication  & 46.7  & 85.0  & 85.0 & 0.716 & 0.801 & 47.0 & 84.7  & 84.7 & 0.736 & 0.787 & 47.9 & 84.9  & 84.9 & 0.726 & 0.794\\
  PoE  & - & -  & - & - & - & 47.2 & 84.3 & 84.3  & 0.749 & 0.786 & - & - & -  & - & - \\
 Tensor Fusion \cite{Zadeh2017Tensor} & 46.9 & 85.0 & 85.0 & 0.727 & 0.798 & 46.7 & 85.0 & 85.0 & 0.714 & 0.795  & 46.7 & 86.0 & 85.9 & 0.713 & 0.799 \\
 Graph Fusion \cite{ARGF} & 45.8 & 85.2 & 85.2 & 0.726  & 0.783 & 46.9 & 84.3 & 84.3 &  0.732 & 0.791 & 47.1 & 85.5  & 85.4 & 0.716  &   0.794 \\
 \hline
Average & $\underline{47.1}$ & $\underline{85.0}$ & $\underline{85.0}$ & \textbf{0.719}  & \textbf{0.796} & 46.5 & 84.6 & 84.6 &  0.731 & 0.790 & \textbf{47.6} & \textbf{85.2} & \textbf{85.2} &  $\underline{0.721}$ &  $\underline{0.795}$  \\
 \hline
 \end{tabular}}
 \vspace{-0.2cm}
\end{table*}%

\begin{table*}[t]
\centering
 \caption{ \label{t449}\textbf{ Discussion on the fusion strategies  and the MIB variants on CMU-MOSEI dataset.}
 }
\resizebox{1.95\columnwidth}{!}{\begin{tabular}{c|c|c|c|c|c|c|c|c|c|c|c|c|c|c|c}
 \hline
     &  \multicolumn{5}{c|}{E-MIB}   &   \multicolumn{5}{c|}{L-MIB}  &  \multicolumn{5}{c}{C-MIB}  \\
 \hline
     & Acc7 ($\uparrow$) & Acc2 ($\uparrow$) & F1 ($\uparrow$) & MAE  ($\downarrow$) & Corr ($\uparrow$)   & Acc7 ($\uparrow$) & Acc2 ($\uparrow$) & F1 ($\uparrow$) & MAE  ($\downarrow$) & Corr ($\uparrow$)   & Acc7 ($\uparrow$) & Acc2 ($\uparrow$) & F1 ($\uparrow$) & MAE  ($\downarrow$) & Corr ($\uparrow$) \\
 \hline
 Concatenation &  53.2 & 85.4 & 85.4 & 0.588 & 0.790 & 54.1  & 85.7 & 85.7 & 0.600 & 0.788 & 53.0 & 86.2 & 86.2 & 0.584 &  0.789 \\
 Addition  & 54.8  & 85.5 & 85.5 & 0.600 & 0.784 & 53.9  & 85.3 & 85.3 & 0.592 & 0.788 & 54.0 & 85.7 & 85.7 & 0.600 & 0.782   \\
  Multiplication  &  52.6 & 85.4 & 85.4 & 0.606 & 0.774 & 52.8  & 85.1 & 85.2 & 0.612 & 0.775 & 53.0 & 85.5 & 85.4 & 0.596 &  0.786 \\
  PoE  & - & -  & - & - & - & 53.2 & 85.2 &  85.3 & 0.601& 0.782 & - & - & -  & - & - \\
 Tensor Fusion \cite{Zadeh2017Tensor} &  53.7 & 85.4 &  85.5 & 0.605 & 0.776 & 53.6  & 85.3 & 85.3 & 0.604 & 0.777 & 54.3 & 86.3 & 86.4 & 0.597 &  0.787  \\
 Graph Fusion \cite{ARGF} & 54.6  & 85.5 & 85.3 & 0.590 & 0.786 &  53.5 & 85.4 & 85.4 & 0.598 & 0.779 & 54.5 & 86.3 & 86.2 & 0.590 &  0.790  \\
 \hline

Average &  \textbf{53.8} & $\underline{85.4}$ & $\underline{85.4}$ & $\underline{0.598}$ & $\underline{0.782}$ & $\underline{53.5}$  & 85.3 & $\underline{85.4} $& 0.601 & $\underline{0.782}$ & \textbf{53.8} & \textbf{86.0} & \textbf{86.0} & \textbf{0.593} & \textbf{0.787}     \\
 \hline
 \end{tabular}}
 \vspace{-0.3cm}
\end{table*}%

\subsection{\textbf{Discussion on Fusion Architectures and the MIB Variants}}\label{sec:fusion_e}

In this section, we comprehensively evaluate and compare the performance of the MIB variants by incorporating various fusion strategies. Moreover, we verify that the our MIB variants are compatible with various fusion strategies and can provide higher expressive power. Analyzing the results on Table~\ref{t4} and ~\ref{t449}, we have the following observations:

1) \textbf{The comparison between the MIB variants:} We use the average results of all the fusion methods as the metrics for a more comprehensive comparison between the MIB variants. Specifically, C-MIB reaches the best performance on the majority of the average metrics on both datasets, while E-MIB reaches the best score on some of the average metrics. Combining all the metrics, C-MIB slightly outperforms E-MIB. L-MIB is the worst version among all the MIB variants. This is reasonable because the multimodal representation is directly used for prediction, and one should directly add the minimal information constraint on the multimodal representation to obtain a better performance. However, L-MIB only adds the constraints to the unimodal representations such that the generated multimodal representation might still be redundant (especially when high-dimensional multimodal representation is generated). Therefore, L-MIB only achieves a sub-optimal solution. Notably, the differences in the average performance between the MIB variants are consistent across two datasets.

Interestingly, we observe that when the complex fusion methods (tensor fusion and graph fusion) are applied, the performance of the C-MIB improves considerably. In contrast, the performances of the L-MIB and the E-MIB are not significantly improved. This is because E-MIB cannot explicitly filter out the noise in unimodal representations, and the noise may seriously influence the quality of  the multimodal representation when complex fusion mechanisms are applied. For the L-MIB, it cannot reduce the redundancy of the multimodal representation. Actually, when complex fusion mechanisms are used, the generated multimodal representation is often highly complicated and  redundant. Therefore, the performance of L-MIB is inferior to C-MIB. 

2) \textbf{The comparison with PoE:}  By comparing the performance of PoE \cite{poe}, we demonstrate the advantage of our MIB over the IB methods applying on multi-view learning \cite{mvib,mvib2,MVAE}. PoE \cite{poe} is similar to weighted average and is widely-used in many multi-view IB methods \cite{poe,mvib,mvib2,MVAE}. However, it performs unfavorably compared to other fusion methods injecting into L-MIB. It demonstrates that we should enable the framework to be compatible with other fusion methods such as tensor fusion to provide higher expressive power. The compatibility with other fusion methods is one of the advantages of our MIB framework but cannot be realized by previous multi-view IB methods \cite{poe,mvib,mvib2,MVAE}.

3) \textbf{The comparison between  fusion methods:} Concatenation, Addition, and Multiplication are the simple fusion methods, which cannot explore complex inter-modal dynamics. Compared with them, more complicated fusion strategies such as graph fusion and tensor fusion can further learn complex cross-modal dynamics that benefit prediction. However, they face a higher risk of generating redundant information and increasing the computational cost. In general, Concatenation outperforms the other simple fusion methods on most of the evaluation metrics across two datasets and three MIB variants. This is reasonable because Concatenation can well-preserve the unimodal information when generating the multimodal representation, which cannot be realized by Addition and Multiplication.
In addition, combining the results from the two datasets, tensor fusion reaches the best performance among all the fusion methods, which slightly outperforms graph fusion. We argue that this is because tensor fusion can generate high-expressive and high-dimensional multimodal tensor to explore the complicated interactions between modalities. And via the minimal information constraint of our MIB, the redundancy and the noise of unimodal representations and the multimodal tensor are reduced, such that the generated multimodal tensor is of high quality and suitable for prediction.

\begin{table}[t]
\centering
 \caption{ \label{t334}\textbf{Discussion of unimodal and bimodal systems on the CMU-MOSI dataset.} `L', `A', and `V' denote language, acoustic, and visual modality, respectively. }
\resizebox{.95\columnwidth}{!}{\begin{tabular}{c|c|c|c|c|c}
 \hline
     & Acc7 ($\uparrow$) & Acc2 ($\uparrow$) & F1 ($\uparrow$) & MAE  ($\downarrow$) & Corr ($\uparrow$) \\
 \hline
L   &  \underline{46.6} &  83.5 &  83.5 & 0.728 & 0.792  \\
L (W/O IB)  &  46.0 &  83.1 & 83.0  & 0.731 &  0.795 \\
\hline
 A   &  18.4 & 55.7  &  54.2 & 1.478 &  0.126 \\
 A (W/O IB)  &  16.2 & 55.1  & 53.7  & 1.474 &  0.007 \\
\hline
 V  &  21.8 & 57.1  & 57.0  & 1.460 & 0.060  \\
 V (W/O IB)  & 18.1  &  53.9 & 54.2  & 1.477 &  0.059 \\
 \hline
L \& A   &  45.7 & \textbf{84.4}  &  \textbf{84.4} & \textbf{0.713} & \textbf{0.801}  \\
L \& A (W/O IB)  &  46.3 & 83.4 & 83.4  & 0.734 & 0.794  \\
\hline
 A \& V  & 20.3  & 58.6  & 57.4  & 1.445 &  0.165 \\
  A \& V (W/O IB) &  18.7 &  54.1 & 53.9  & 1.490 &  0.025 \\
  \hline
L  \& V  &  $\underline{46.6}$ & $\underline{84.3}  $&  $\underline{84.3}$ & $\underline{0.717}$ & $ \underline{0.800}$ \\
L  \& V (W/O IB) &  \textbf{47.1} & 83.8  & 83.8  & 0.727 & 0.791   \\
  \hline
 \end{tabular}}
 \vspace{-0.3cm}
\end{table}%

\subsection{\textbf{Analysis on Unimodal and Bimodal Systems}}
In this section, we provide the results of the unimodal and bimodal systems as well as their counterparts where the IB regularization is removed. By observing the results in Table~\ref{t334}, we have the following conclusions:

\textbf{1) The importance of each modality: } The language modality, with the powerful BERT \cite{BERT} framework as its feature extractor, significantly performs better than the visual and acoustic modalities. Moreover, visual modality performs better than acoustic modality. However, the performances of these two modalities are actually unsatisfactory. One of the reasons is that the extracted features of these two modalities are of low quality and not sentiment-sensitive. This highlights a possible direction for improving the performance of the multimodal sentiment analysis system: extracting high-quality features of the nonverbal modalities for sentiment analysis.

\textbf{2) The discussion on bimodal systems:}
Generally, the bimodal systems outperform their unimodal counterparts, and the trimodal system performs better than its bimodal counterparts (see Table~\ref{t1} for the results of trimodal system). These results demonstrate that integrating more modalities helps improve the performance. Notably, even with a lowest performance on acoustic modality, the language-acoustic system achieves the best performance among the bimodal systems. The result indicates that language and acoustic modalities have more complementary information between each other.

\textbf{3) The effectiveness of the IB principle on unimodal and bimodal systems:}
As shown in Table~\ref{t334}, adding the IB constraints consistently improves the overall performance. Moreover, we observe that the IB principle is more beneficial to those modalities with more noises: the improvement of the IB regularization is more significant on visual, acoustic, and visual-acoustic systems. These results to some extent demonstrate that the IB regularization helps reduce the noise in the features and improve the performance.

\begin{table}[t]
\centering
 \caption{ \label{t333}\textbf{ Analysis on the Hyper-parameter $\beta$.} }
\resizebox{.95\columnwidth}{!}{\begin{tabular}{c|c|c|c|c|c}
 \hline
     & Acc7 ($\uparrow$) & Acc2 ($\uparrow$) & F1 ($\uparrow$) & MAE  ($\downarrow$) & Corr ($\uparrow$) \\
 \hline
 $\beta$ =1e0  & 38.0 &  80.5 &  80.5 & 0.872 & 0.724 \\
$\beta$ =1e-1  & 47.2 & 84.4 & 84.4 & $\underline{0.718}$ & $\underline{0.797} $ \\
  $\beta$ =1e-2 & 46.7 & 84.9 & 84.9 & 0.731 &  0.793   \\
  $\beta$ =1e-3  & \textbf{48.2}  & $\underline{85.2}$  &  $\underline{85.2}$ &  0.728 & 0.793 \\
$\beta$ =1e-4  & 47.6 & 84.7  & 84.7  & 0.728 & 0.793 \\
$\beta$ =1e-5  & $\underline{48.0}$ & \textbf{85.7}  &  \textbf{85.6} & \textbf{0.709} & \textbf{0.799} \\
$\beta$ =1e-6  & 46.6 & 84.3  &  84.3 & 0.727 & 0.795 \\
$\beta$ =0  & 46.3 &  83.8 & 83.8  & 0.730 & 0.790 \\
  \hline
 \end{tabular}}
 \vspace{-0.3cm}
\end{table}%

\subsection{\textbf{Analysis on the Hyperparameter $\beta$}}
In this section, we conduct analysis on the hyperparameter $\beta$ which controls the strength of the minimal information constraint (see Eq.~\ref{C-MIB_t}). $\beta$ also balances the non-discriminative and discriminative information in the latent representations. The experiment is conducted on C-MIB. From the results in Table~\ref{t333}, we can observe that when $\beta$ becomes larger ($\beta$=1), the performance drops significantly.
Notably, when $\beta$ is equal to zero, binary accuracy drops by 1.5\% and  7-class accuracy  drops by 2\% compared to the performance when $\beta$ is 1e-3, and the MAE and Corr also become worse. Actually, in the case that $\beta$ is zero, the minimal information constraint is removed, and the framework degenerates into a regular framework whose objective is maximizing the mutual information between representations and the target. The decline on performance demonstrates that the introduced IB principle is effective and can boost the performance. We argue that this is because the minimal information constraint helps to filter out the noise and reduce the redundancy of unimodal and multimodal representations.


Actually, the model performs best when $\beta$ is set to 1e-5. To be consistent with the setting of L-MIB and E-MIB, the defaulted value of $\beta$ is set to 1e-3. All in all, these results indicate that $\beta$ should be a small number such that the minimal information constraint will not dominate the learning and more information can be retained, which prevents the IB principle from filtering out the necessary information. Meanwhile, it is necessary to reduce the redundancy and filter out the noise, as the performance drops considerably when the minimal information constraint is removed.

\begin{figure}[htbp]
 \vspace{-0.3cm}
\centering
\subfigure[Learned without C-MIB]{
\begin{minipage}[t]{0.45\linewidth}
\centering
\includegraphics[width=1.5in]{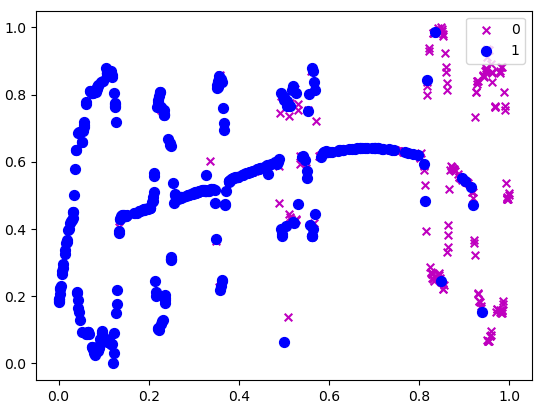}
\vspace{-0.3cm}
\end{minipage}%
 }
 \subfigure[Learned by C-MIB]{
 \begin{minipage}[t]{0.45\linewidth}
 \centering
 \includegraphics[width=1.5in]{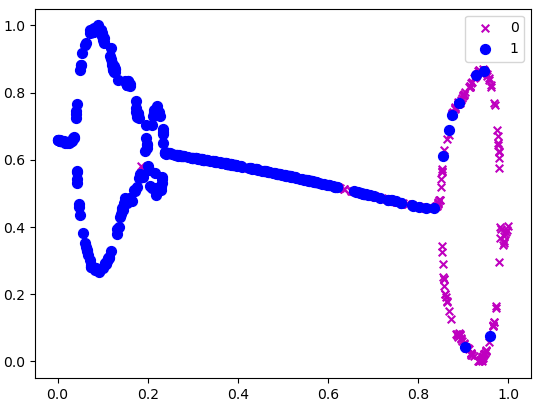}
 \vspace{-0.3cm}
 \end{minipage}
 }%
 \centering
  \vspace{-0.3cm}
 \caption{\label{8}\textbf{Visualization of the multimodal representations on CMU-MOSI testing set.} The `purple x' and `blue dot' denote the data point for positive and negative sentiment, respectively. }
 \vspace{-0.3cm}
 \end{figure}

\subsection{\textbf{Visualization of the Embedding Space}}\label{sec:visual}

In this section, we provide a visualization for distributions of the multimodal representations to analyze the class distributions in the embedding space. Following \cite{ARGF,MISA}, the visualization is obtained by transforming each representation into 2-dimensional feature point using the t-SNE algorithm\cite{tsne}, which can capture much of the local structure of the high-dimensional data very well. The left sub-figure on Fig.~\ref{8} illustrates the embedding space learnt without C-MIB, and the right sub-figure presents the embedding space learnt by the C-MIB.  We can infer from Fig.~\ref{8} that  when the IB principle is not applied, the data points of the multimodal representations in the embedding space are very scattered, and each class (positive or negative class) does not form a distinguishing cluster. In contrast, when the IB principle is applied, the data points from the same class form a discriminative cluster. The distance between the centers of the positive and negative sentiment clusters is far away, and the data points that are difficult to be clustered are placed between the two clusters. The visualization results reveal an important property of our C-MIB, i.e. enabling the multimodal representation to be more discrimiative by reducing the redundancy and filtering out the noise in unimodal and multimodal features.

\subsection{\textbf{Discussion on Minimal Information Constraint}}
The minimal information constraint aims to constrain the mutual information of the encoded representation $\bm{z}$ and the input $\bm{x}$. It encourages $\bm{z}$ to forget the information from $\bm{x}$. In this sense, the minimal information constraint can be replaced by other functions that control the `similarity' between $\bm{z}$ and $\bm{x}$. In this section, we replace the minimal information constraint (KL-divergence term) with other objectives to reveal its significance. The compared objectives include maximum mean discrepancy (MMD), Deep CORAL\cite{sun2016deep}, orthogonal constraint, European distance and cosine distance.  MMD is widely used in domain adaption to measure the distribution discrepancy between the source and target domain. Here we apply MMD to encourage the distribution discrepancy between $\bm{z}$ and $\bm{x}$ to be maximum such that they are not similar in terms of distribution. We use the Gaussian kernel as the kernel function of MMD. Since MMD is widely used, we omit the computation of the MMD loss $l_{MDD}^{'}$ for conciseness. Since we need to maximize MMD, the final MMD loss is calculated as: $l_{MDD}=\frac{1}{1+l_{MDD}^{'}}$ where 1 is used in the denominator to prevent division by zero. Deep CORAL is similar to MMD which adds constraints on the distribution between $\bm{z}$ and $\bm{x}$:
\begin{equation}
  \begin{aligned}
  &C_{x} =\frac{1}{n-1}\left(\bm{x}^T \bm{x}-\frac{1}{n}\left(\mathbf{1}^{T} \bm{x}\right)^{T}\left(\mathbf{1}^{T} \bm{x}\right)\right) \\
  &C_{z} =\frac{1}{n-1}\left(\bm{z}^{T} \bm{z}-\frac{1}{n}\left(\mathbf{1}^{T} \bm{z}\right)^{T}\left(\mathbf{1}^{T} \bm{z}\right)\right) \\
  l_{C O R A L}^{'}&\!=\!\frac{1}{4d^{2}}\left\|C_{S}-C_{T}\right\|_{F}^{2} \ l_{C O R A L}\!=\!\frac{1}{1+l_{C O R A L}^{'}}\\
  \end{aligned}
\end{equation}
where $\mathbf{1}$ is the column vector with all elements equal to 1, $||\cdot||^2_F$ denotes Frobenius norm, $d$ is the feature dimensionality, $n$ is the sampling size,
and $T$ denotes the transpose operation. In comparison, European distance forces the distance between $\bm{z}$ and $\bm{x}$ to be maximum such that they are not similar in terms of representations, and cosine distance likewise explains.
Orthogonal constraint\cite{MISA} forces $\bm{z}$ and $\bm{x}$ to be orthogonal:
\begin{equation}
\begin{aligned}
  \bm{x} \longleftarrow \frac{\bm{x}}{\sqrt{\sum_j \bm{x}_j^2}+1}&, \ \ \bm{z} \longleftarrow \frac{\bm{z}}{\sqrt{\sum_j \bm{z}_j^2}+1} \\
  l_{oc} &= \bm{x}\bm{z}^T
  \end{aligned}
\end{equation}

As presented in Table~\ref{t444}, the original KL-divergence term performs best, and the other methods also reach competitive performance compared to the case that minimal information constraint is removed (see `W/O IB' case on Table~\ref{t1}). These results demonstrate that the idea of controlling the information from the input and filtering out noisy information is effective. Since the KL-divergence constraint explicitly measures the mutual information between the input and the encoded representation, it can reach a better performance compared to other methods that only measure the similarity of distribution or feature embedding between the input and the encoded one.

\begin{table}[t]
\centering
 \caption{ \label{t444}\textbf{ Discussion on Minimal Information Constraint.} }
\resizebox{.95\columnwidth}{!}{\begin{tabular}{c|c|c|c|c|c}
 \hline
 & Acc7 ($\uparrow$) & Acc2 ($\uparrow$) & F1 ($\uparrow$) & MAE  ($\downarrow$) & Corr ($\uparrow$) \\
 \hline
 MMD  & 46.6 &  84.6 & 84.6 & 0.725 & $\underline{0.798}$ \\
  Deep CORAL  & 44.7 &  84.4 & 84.4 & 0.735 & $\underline{0.798}$ \\
 Orthogonal Constraint  & $\underline{47.6}$ &$ \underline{84.9}$  & $\underline{84.9} $& \textbf{0.717} &  $\underline{0.798}$ \\
  European distance  & 45.8 &  84.6 & 84.6 & 0.732 & 0.795 \\
Cosine Distance  & 46.6 &  83.8 & 83.8 & $\underline{0.719}$ & \textbf{0.800} \\
  \hline
C-MIB  & \textbf{48.2}  & \textbf{85.2}  &  \textbf{85.2} &  0.728 & 0.793  \\
 \hline
 \end{tabular}}
 \vspace{-0.2cm}
\end{table}%

\subsection{\textbf{Learning Curve for Minimal Information Constraints}}

\begin{figure}
\setlength{\belowcaptionskip}{-0.3cm}
\centering
\includegraphics[scale=0.19]{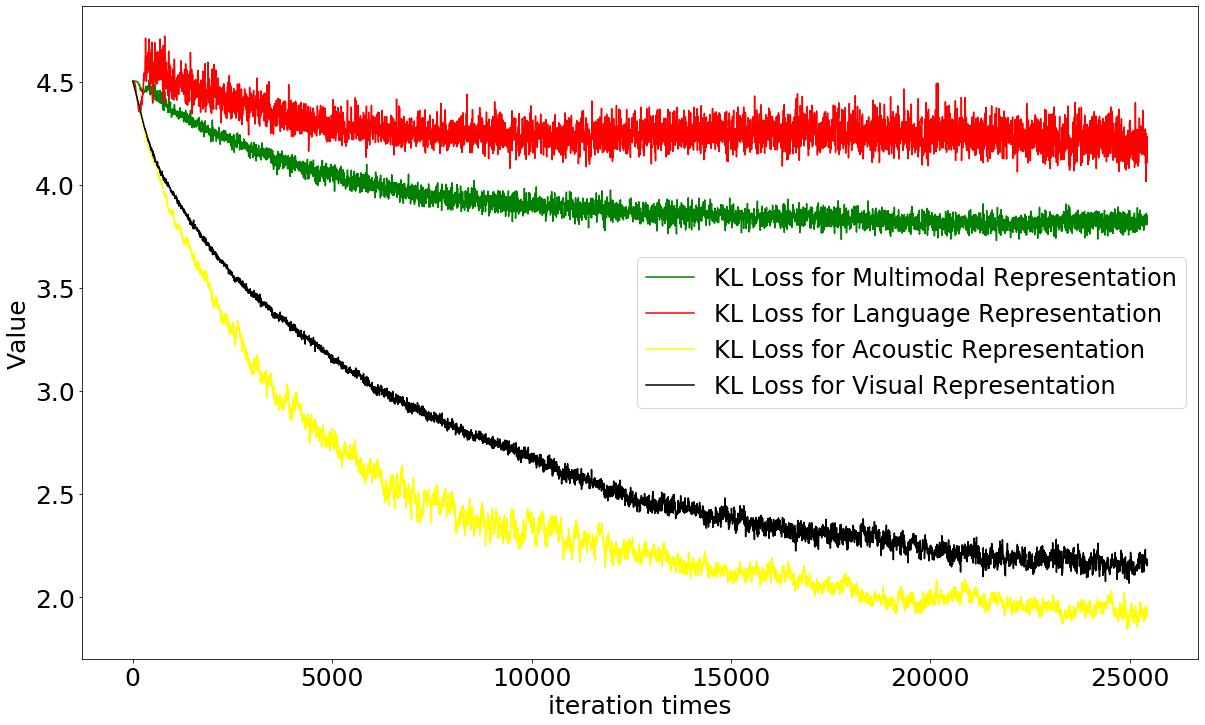}
\caption{\label{visual2}\textbf{Learning Curve for Minimal Information Constraint Losses of C-MIB.} The `KL Loss' in the figure denotes minimal information constraint loss.
}
\vspace{-0.2cm}
\end{figure}

In this section, we present the learning curves for the minimal information constraint losses (i.e., the KL-divergence losses). As shown in Fig.~\ref{visual2}, the losses of acoustic and visual modalities become much lower as the training deepens, indicating that these two modalities contain many noises that need to be filtered out. In contrast, the loss for language modality drops very little, suggesting that language modality contains sufficient discriminative information for predicting the target. These results are consistent with previous work \cite{TASLP,MCTN,HFFN}, and indicate that our method can explicitly filter out noisy information and retain useful information. As for the loss for multimodal representation, it drops moderately and is convergent in the middle of the training, which suggests that the generated multimodal representation still has redundant information and needs to be explicitly handled.

\begin{table}[!htb]
\centering
\caption{ \label{t666}\textbf{The comparison of model complexity.}}
\resizebox{.82\columnwidth}{!}{\begin{tabular}{c|c}
 \hline
    & The number of parameters\\
 \hline
MAG-BERT \cite{MAG-BERT} & 110,853,121\\
Ours (W/O IB) & \textbf{109,216,699}\\
E-MIB & \underline{109,371,423} \\
L-MIB & 109,681,024\\
C-MIB & 109,835,748\\
 \hline
 \end{tabular}}
  \vspace{-0.3cm}
\end{table}%

\subsection{\textbf{Analysis on Model Complexity}}
We analyze the model complexity of the proposed method on CMU-MOSI in this section. Since our MIB variants require additional encoders and decoders to process the signal, the number of parameters increases after the injection of the IB regularization. We can infer from Table~\ref{t666} that when the IB principle is introduced, the number of parameters for the C-MIB is 109,835,748. The number of parameters becomes 109,216,699 when the IB constraints are removed. The total number of parameters introduced by the IB components therefore is 619,049, which is 0.564\% of the total number of model parameters. Therefore, it is not an expensive cost to introduce the IB regularization. Notably, a larger model is not necessarily a better model, because the better-performing E-MIB has fewer parameters than the L-MIB. For comparison, we also present the number of parameters of the state-of-the-art model MAG-BERT \cite{MAG-BERT}. It can be seen that C-MIB has fewer parameters than MAG-BERT but still reaches a better result, which further demonstrates the effectiveness of C-MIB.

\section{Conclusion}\label{sec:Conclusion}

We propose three MIB variants, namely L-MIB, E-MIB, and C-MIB, which focus on different perspectives of multimodal representation learning. MIB  aims to learn minimal sufficient unimodal/multimodal representation, and is compatible with any fusion mechanisms which enables flexibility. Our method achieves state-of-the-art performance on two multimodal tasks. Moreover, the visualization suggests that the IB principle helps find more discriminative multimodal representation. We also use various objectives to replace the minimal information constraint in the MIB, all of which achieve promising result.

\ifCLASSOPTIONcaptionsoff
  \newpage
\fi

\bibliographystyle{IEEEtran}


%

%

\bibliography{sentiment2}

\end{document}